\documentclass{article}
\usepackage{microtype}
\usepackage{spconf,amsmath,graphicx}
\usepackage{xcolor}
\usepackage{balance}
\usepackage{hyperref}

\usepackage[ruled,vlined]{algorithm2e}

\usepackage{multirow}

\title{Automatic Cross-Domain Transfer Learning for Linear Regression}
\name{Liu Xinshun, He Xin, Mao Hui, Liu Jing, Lai Weizhong, Ye Qingwen}
\secondlinename{liuxinshun12@163.com}
\address{Vivo Software Co. LTD, Beijing, China}

\begin{document}
%
\maketitle
\begin{abstract}
Transfer learning research attempts to make model induction transferable across different domains. This method assumes that specific information regarding to which domain each instance belongs is known. This paper helps to extend the capability of transfer learning for linear regression problems to situations where the domain information is uncertain or unknown; in fact, the framework can be extended to classification problems. For normal datasets, we assume that some latent domain information is available for transfer learning. The instances in each domain can be inferred by different parameters. We obtain this domain information from the distribution of the regression coefficients corresponding to the explanatory variable $x$ as well as the response variable $y$ based on a Dirichlet process, which is more reasonable. As a result, we transfer not only variable $x$ as usual but also variable $y$, which is challenging since the testing data have no response value. Previous work mainly overcomes the problem via pseudo-labelling based on transductive learning, which introduces serious bias. We provide a novel framework for analysing the problem and considering this general situation: the joint distribution of variable $x$ and variable $y$. Furthermore, our method controls the  bias well compared with previous work. We perform linear regression on the new feature space that consists of different latent domains and the target domain, which is from the testing data. The experimental results show that the proposed model performs well on real datasets.

\end{abstract}

\section{Introduction}

In the past few decades, transfer learning has received substantial interest. Transfer learning supposes that there are domain concepts for data; typically, the training data is considered as the source domain, and the testing data is considered as the target domain. Furthermore, because different situations can occur, there can be multiple source domains; such problems are regarded as multitask learning problems. Transfer learning is applied when distributions differ by domain but the domains are closely related. Learning multidomain cases simultaneously may lead to better generalizability than  taking all domains as a whole. However, for the above transfer learning methods, the domain information must be known. Unfortunately, such information is not always available, and the group strategy is not explicit. In this case, we may take the entire training dataset to perform traditional machine learning or transfer learning. In this paper, we propose an automatic cross-domain transfer learning method to first identify the latent domain information and then perform a transfer process based on the learned domain information to improve the learning performance. The proposed framework uses a Dirichlet process technique to mine the latent domain information. This technique can create clusters according to regression coefficients, which is reasonable for our latent domain mining task. Based on this technique, we characterize the domain relation according to the Dirichlet distribution. Furthermore, we need to transfer the latent domains and the target domain, which is obtained from the testing data. Since our latent domains are mined according to the regression coefficients, which correspond to the explanatory variable $x$ and the response variable $y$, our transfer process should consider the joint distribution $p(x,y)$. However, response value $y$ is not available for the testing data. Previous work mainly handles this problem via pseudo-labelling according to transductive learning, which would seriously bias our framework since the pseudo-label is quite inaccurate without considering latent domains. Thus, we propose a novel framework for analysing the problem, and our method efficiently controls the bias. Through the learned feature space, the knowledge of one latent domain or target domain can be transferred to another more accurately than previous work. Then, this framework can be used to dramatically extend the transfer learning method, as only traditional machine leaning models that consider the entire training dataset can be applied in this scenario. 

This paper considers a linear regression problem based on the training dataset $\{(x_i, y_i )\}_n$ with $n$ instances. The linear regression problem corresponds to explanatory variable $x_i$ and response value $y_i$, where $x_i$ has $p$ dimensions. 

Note that in our framework, the domains in the training data are latent. As a result, we must define the concept of a latent domain. Traditionally, different domains should have different coefficients. A study by~\cite{ref_article1} suggests assigning each sample its own proprietary regression coefficient and applying a Dirichlet process prior to these regression coefficients for regularization; ~\cite{ref_article1} considers multivariate regression, and the corresponding coefficient is a matrix. However, our corresponding coefficient is a vector. The Dirichlet process clustering property helps to capture the latent domain information for our framework, and the Polya urn scheme for Bayesian inference in linear regression is employed in the feature space. Furthermore, the linear regression model is a conjugate model that can be inferred via a relatively straightforward use of Markov chain Monte Carlo techniques. 

As demonstrated above, after mining latent information, we need to transfer the learned domain and the target domain. However, considering only the distribution of $x$ may not work, as our latent domain is mined according to the coefficients of the linear regression model, which corresponds to the response variable $y$. Then, we try to transfer the joint distribution $p(x,y)$. Figure 1 presents the process applied in our work. The first picture shows the original linear regression problem. The stars represent the testing data. We mine the latent domain via the Dirichlet process for the training data, as shown in the second picture. The third picture shows the transfer results for the distribution $x$, and the fourth picture shows the transfer results for the joint distribution $p(x,y)$. Compared with the third picture, the fourth picture shows the successful transfer of the latent domains. Since the latent domains are mined via the regression coefficients, they cannot be discriminated merely on explanatory variable $x$.

As in the considered scenario, there are no response values $y$ for the testing data. Our framework attempts to transfer the latent domains via explanatory variable $x$, response value $y$, and the target domain with only explanatory variable $x$. Some studies consider the problem by using seed labels~\cite{ref_article29} or pseudo-labels via transductive learning~\cite{ref_article3}~\cite{ref_article23}; however, the first method requires prior information, and the second method introduces a serious bias. For our problem, since we consider a method with multiple latent domains, which is quite complex, the pseudo-labels would be more inaccurate. In our framework, the target domain lacks the dimension of $y$, which is equivalent to taking the response variable $y$ as zero. Moreover, the response variable $y$ for the training data can be normalized via the z-score and reduced to simplify the transfer between the latent domains and the target domain. Furthermore, in our framework, the bias of the response variable $y$ for the testing data can also be controlled efficiently via regularization. In this way, we can learn a new feature space such that the joint distributions of the related latent domains and the target domain are close to each other. Thus, knowledge of the latent domains can be transferred. In this paper, we propose a novel method for adapting the joint distribution $p(x,y)$ to overcome the defects that the testing data have no response variable $y$. In contrast to previous work, our method does not need prior information and can efficiently control the introduced bias as demonstrated before. 

\begin{figure}[!htb]
	\centering
	\includegraphics[width=3.9cm,height=3cm]{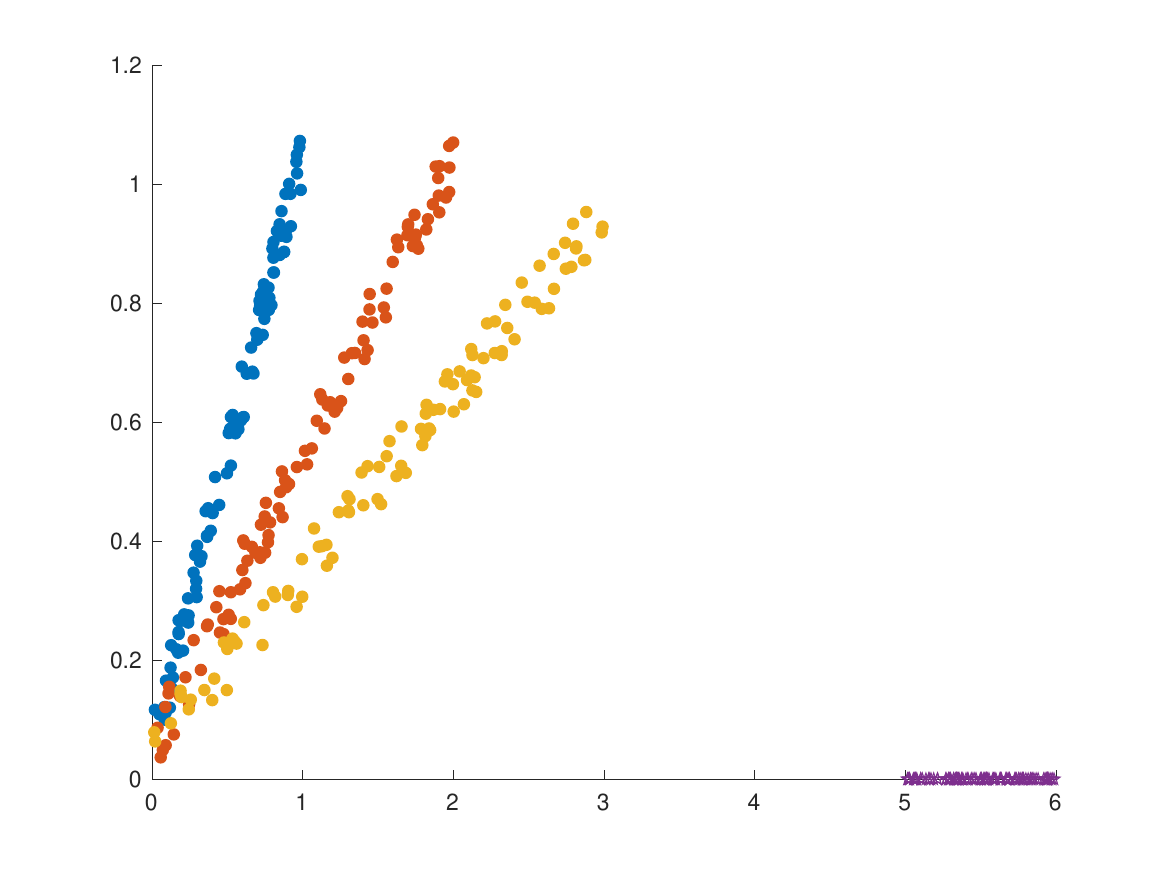}
	\includegraphics[width=3.9cm,height=3cm]{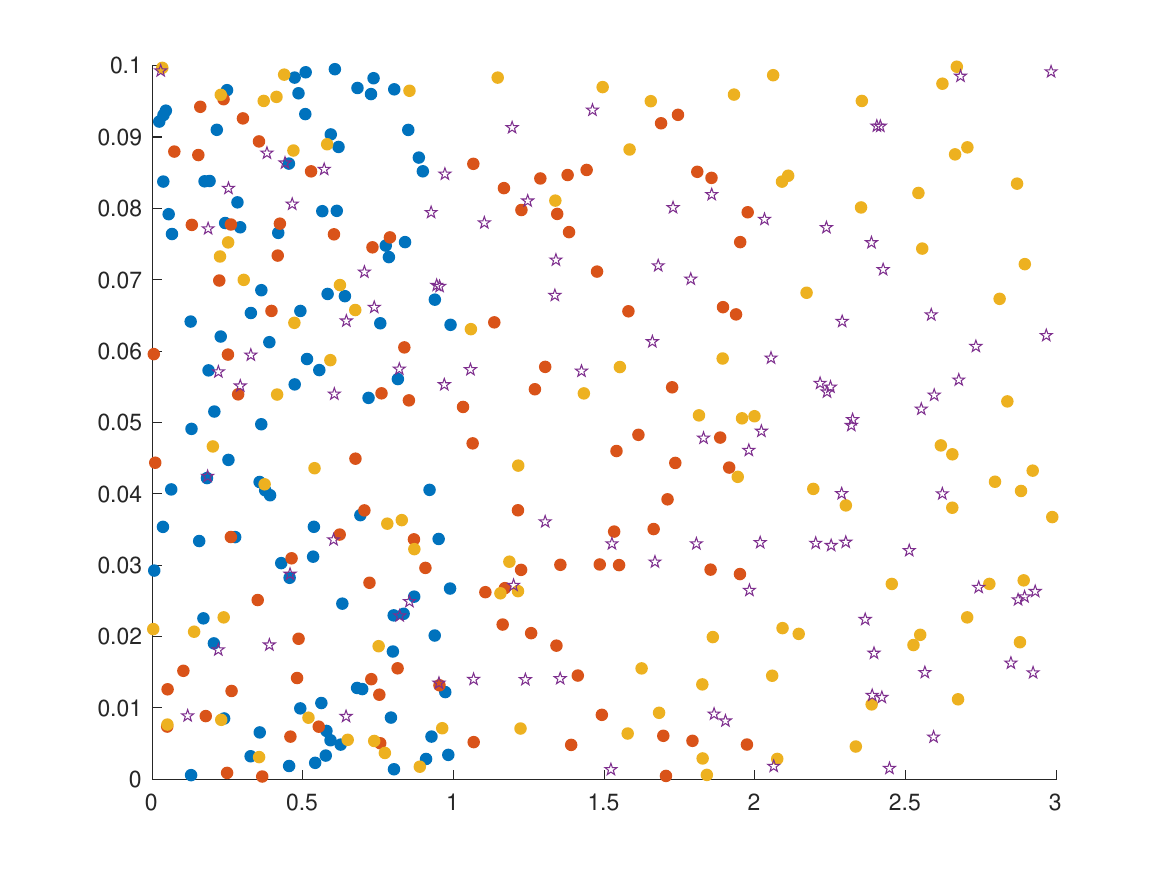}
	\includegraphics[width=3.9cm,height=3cm]{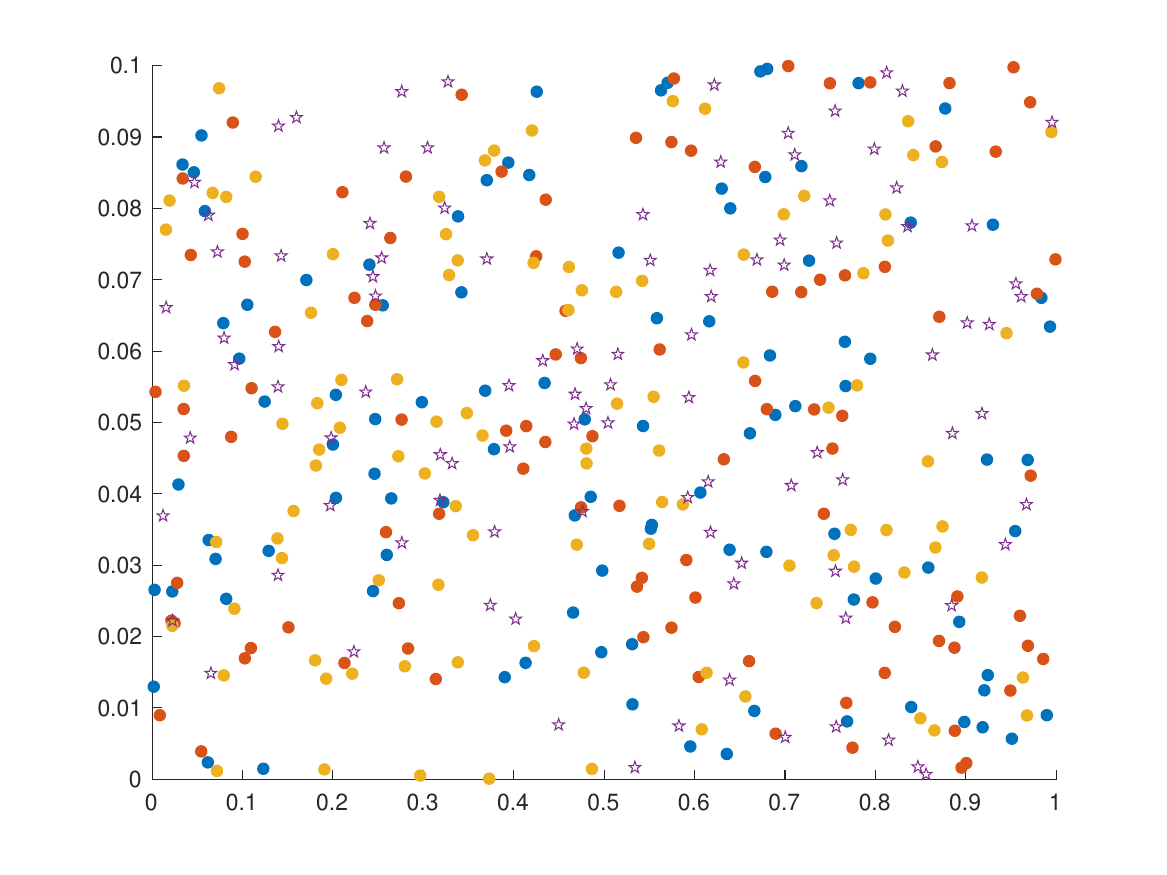}
	\caption{The first picture shows the original linear regression problem. The star points represent the testing data. The second picture shows the latent domains mined via the Dirichlet process for the training data. The third picture shows the transferred results based on the distribution of $x$, and the fourth picture shows the transferred results of the joint distribution $p(x,y)$. Compared with the third picture, the fourth picture shows the successful transfer of the latent domains.}
	\label{fig2}
\end{figure}

Furthermore, previous studies ~\cite{ref_article2}~\cite{ref_article3}~\cite{ref_article4} suggest regularizing the transfer process using PCA or the Laplace graph to preserve the main feature vector or the manifold of the domains; we neatly integrate these processes and our proposed regularization for response variable $y$ in the framework. Finally, we apply our framework to real datasets to demonstrate that the proposed method can extend transfer learning dramatically and achieve better performance.

This paper makes three main contributions:

1. We extend transfer learning to situations in which domain information is uncertain. Our method focuses on latent domains. This would help to get better performance than taking the traing data as a whole which is the benefit of transfer learning and it is also efficiently verified in our experiments.

2. We propose a novel method for adapting $p(x,y)$ between latent domains and the target domain, which is from the testing data in our framework. In contrast with previous work, our method does not require prior information and can efficiently control the bias associated with an unknown response variable $y$ for the testing data.

3. We neatly integrate PCA regularization, which is effective and robust in situations with substantial distribution differences, graph regularization, which smooths the graph with respect to the intrinsic manifold structure of the original data, and response regularization, which can control the bias associated with an unknown response variable $y$ for testing data, in one formula.

\section{Related work}
Traditional methods for characterizing domain relations include learning a shared subspace~\cite{ref_article5}~\cite{ref_article6}~\cite{ref_article8}, using a common prior of model parameters~\cite{ref_article1}~\cite{ref_article9}~\cite{ref_article10}, kernel methods~\cite{ref_article11}~\cite{ref_article12}~\cite{ref_article13}~\cite{ref_article14}, Markov logic networks, parameter learning and Bayesian network structure learning~\cite{ref_article15}. However, these methods assume specific domain information, such as to which domain each instance belongs. Our framework extends the ability of transfer learning to the scenario in which the domain information is uncertain.

Some similar works use mixture models and group instances for regression or classification~\cite{ref_article1}~\cite{ref_article16}~\cite{ref_article17}~\cite{ref_article18}, similar to how we attempt to mine the latent domain information. Furthermore, our work is related to dictionary methods and sparse coding~\cite{ref_article19}~\cite{ref_article20}~\cite{ref_article21}. The instances in the same latent domain tend to have the same code. However, all these works attempt to group instances and obtain the distribution. New data are handled from a maximum likelihood perspective, and the relationships among the latent domains and their transfer effects are not considered. Our work focuses on transfer learning to identify the relationships among of the latent domains and the target domain and then achieve better performance.

Deep learning has achieved good performance in many research fields. Deep transfer learning studies how to utilize knowledge from other fields by means of deep neural networks~\cite{ref_article22}~\cite{ref_article23}~\cite{ref_article24}~\cite{ref_article25}~\cite{ref_article26}~\cite{ref_article27}~\cite{ref_article28}. However, methods such as DANN also needs to group information and transfers only the distribution of $x$. By contrast, we consider a better joint distribution $p(x,y)$ as demonstrated previously. 

Other studies have considered transferring the response variable $y$ of testing data by using either seed labels~\cite{ref_article29} or pseudo-labels~\cite{ref_article3}~\cite{ref_article23} via transductive learning. However, the first method requires prior information, and the second method introduces serious bias. In contrast, our framework does not require prior information and can control the bias efficiently.

\section{Proposed Framework}
First in this paper, we focus on the following linear regression model \begin{equation}y_i = x_i\mathcal{A}_i + \epsilon_i,\end{equation} where $x_i=(1,x_{i1},x_{i2}, ..., x_{ip})$,  $\mathcal{A}_i$ is the $(p+1) \times 1$ coefficient vector and $\epsilon_i$ is the residual error. Note that we first suppose that each sample has its own coefficient. Then, we formulate our framework as follows.

\subsection{Latent Domain Information Based on a Dirichlet Process}
Traditionally, domain information would be characterized by the coefficients; that is, instances in the same domain have the same regression coefficients, and instances in different domains have different coefficients. According to~\cite{ref_article1}, a Dirichlet process is considered to be the prior distribution of $\mathcal{A}_i$. The clustering property of the Dirichlet process helps us to find the latent domain. However, ~\cite{ref_article1} consider the process for multivariate regression, where $\mathcal{A}_i$ is a matrix, and in our framework, $\mathcal{A}_i$ is a vector. Here, we take a Dirichlet process as the prior under the assumption that the clustering property would formulate the latent domain in a natural manner. Suppose we have $n$ instances for training data. The response value for the $n$ instances follows a Dirichlet process mixture model:

\begin{align*}
[y_i|\mathcal{A}_i,\Sigma] &\sim N(y_i|x_i\mathcal{A}_i, \Sigma), i = 1,...,n; \\
[\mathcal{A}_i|G] &\sim G, i = 1,...,n; \\
G &\sim DP(\nu G_0).
\end{align*}
This mixture module contains three parameters: the variance $\Sigma$, the concentration parameter of the Dirichlet process prior $\nu > 0$ and the base distribution $G_0$, which can be defined as a normal distribution: \begin{equation}G_0(.|\Sigma, \Lambda) \sim N_{p+1}(0, \Lambda \otimes \Sigma).\end{equation} Note that the covariance matrix of $G_0$ can be a tensor of parameter $\Sigma$. Here, parameter $\Lambda=diag(\lambda_1,\lambda_2, ..., \lambda_{p+1})$ is a diagonal matrix with $\lambda_i > 0$ for $i=1,2, ..., p+1$. Integrating over $G$ for $\mathcal{A}_i$ yields \begin{equation}[\mathcal{A}_i|\mathcal{A}_{-i}] \sim \frac{\nu N_{p+1}(\mathcal{A}_i|0, \Lambda \otimes \Sigma) + \sum_{l \neq i} \delta(\mathcal{A}_i|\mathcal{A}_l)}{\nu + n - 1}.\end{equation} Dirichlet processes have a clustering property that is critical for our latent domain information mining task. Suppose there are $m$ distinct values among $\mathcal{A}_i$ as $Q = \{Q_1, Q_2, ..., Q_m\}$, and $n_k$ contains the occurrences of $Q_k, 1 \leq k \leq m$. 

For our linear regression problem, we can exploit the simple structure of the conditional posterior for each $\mathcal{A}_i$ as follows: for $i = 1,...,n$, the conditional distribution is given by
$$
\begin{aligned}
&[\mathcal{A}_i|\mathcal{A}_{-i},Y,X,\nu,\Lambda,\Sigma] \\
& \propto q_0N(y_i|x_i\mathcal{A}_i,\Sigma)N(\mathcal{A}_i|0,\Lambda \otimes \Sigma) + \sum_{j \neq i}q_j\delta(\mathcal{A}_i|\mathcal{A}_j), \\
\end{aligned}
$$

where

$$
\begin{aligned}
&q_0 = \nu \int N(y_i|x_i\mathcal{A}_i, \Sigma)N(\mathcal{A}_i|0,\Lambda \otimes \Sigma)d\mathcal{A}_i \\
& = \nu N(y_i|0, (x_i \Lambda x'_i + 1)\Sigma),\\
\end{aligned}
$$

and

$$
\begin{aligned}
&q_j = N(y_i|x_i\mathcal{A}_j, \Sigma).\\
\end{aligned}
$$

Integrating over $\mathcal{A}_i$ yields

$$
\begin{aligned}
&[\mathcal{A}_i|\mathcal{A}_{-i},Y,X,\nu,\Lambda,\Sigma] \\
& \propto q_0N(\mathcal{A}_i|C_ix'_iy_i, C_i \otimes \Sigma) + \sum_{k=1}^{m}{n_{k(-i)}q_k\delta(\mathcal{A}_i|Q_k)}, \\
\end{aligned}
$$

where $C_i = (\Lambda^{-1} + x'_ix_i)^{-1}$. Thus, given $Q_k$, with probability $n_{k(-i)}q_k$, we draw $\mathcal{A}_i$ from distribution $\delta(.|Q_k)$, or with probability $q_0$, we draw $\mathcal{A}_i$ from $N(.|C_ix'_iy_i, C_i \otimes \Sigma)$.

The clustering property of Dirichlet processes benefits latent domain information mining. During this process, we also need to resample $Q_k$ after every step. For each $k = 1, ..., m$, we have

$$
\begin{aligned}
&[Q_k|Y,X,\Lambda,\Sigma] \\
& \propto N(Q_k|0, \Lambda \otimes \Sigma) \prod{N(y_i|Q_kx_i, \Sigma)}. \\
\end{aligned}
$$
Then, resampling $Q_k$ can be simplified as \begin{equation}[Q_k|Y, X, \Sigma, \Lambda] \sim N_{p+1}(Q_k|\Theta_kX'_kY_k, \Theta_k \otimes \Sigma),\end{equation} where $Y_k$ and $X_k$ are, respectively, $n_k \times 1$ and $n_k \times (p+1)$ matrices consisting of $y_i$ and $x_i$ with instances belonging to the $k$th cluster and $\Theta_k=(\Lambda^{-1} + X'_kX_k)^{-1}$ for each $k = 1,...,m$. 

Furthermore, the hyperparameters follow their conjugate distributions. Here, $\nu$, $\lambda^{-1}$ and $\Sigma^{-1}$ have Gamma distributions: $Ga(\nu|a_v,b_v)$, $Ga(\lambda_i^{-1}|\frac{a_i}{2},\frac{b_i}{2})$ and $Ga(\Sigma^{-1}|a_0,b_0)$. Then these hyperparameters can be inferred as follow. Update $\Sigma^{-1}$ from $[\Sigma^{-1}|Y,X,\mathcal{A},a_0,b_0]$. \begin{equation}[\Sigma^{-1}|Y,X,\mathcal{A},a_0,b_0] \sim Ga(a_0 + \frac{n}{2}, b_0 + \frac{1}{2}\sum_{i=1}^n(y_i-x_i\mathcal{A}_i)^2)\end{equation} for $i=1,...,n$. Update $\lambda_i^{-1}$ from $[\lambda_i^{-1}|r_i^{k},\Sigma,a_i,b_i]$. \begin{equation} [\lambda_i^{-1}|r_i^{k},\Sigma,a_i,b_i] \sim Ga(\frac{a_i+m}{2}, \frac{b_i+\sum_{k=1}^mr_i^{k}\Sigma^{-1}(r_i^{k})'}{2}),\end{equation} where $r_i^k$ is the $i$th row of $Q_k$. Update $\nu$ from $[\nu|h, a_v,b_v,m]$.
$$
\begin{aligned}
&[\nu|h,a_v,b_v,m] \sim \pi_0Ga(a_v+m,b_v-log(h)) \\
& +(1-\pi_0)Ga(a_v+m-1,b_v-log(h)), \\
\end{aligned}
$$
Here, $h$ is drawn from Beta distribution $Be(\nu+1,n)$, and $\pi_0=\frac{\nu+m-1}{a_v+m-1+n(b_v-log(h))}$.

As mentioned, our latent domain information can be mined by this Dirichlet process, and the learning process is a Gibbs Sampling, as shown in Algorithm 2. Then, we will implement transfer learning in our framework.

\subsection{Absolute Transfer}
Maximum mean discrepancy (MMD), which attempts to evaluate the differences in distributions between domains given finite samples, can be used to compare distributions based on the new transformed space distance. As demonstrated, we use the coefficient parameter to mine the latent domain information that is relevant to the response variable $y$. As a result, we should consider both $x$ and $y$ in the transfer process. However, most traditional transfer methods focus on the distribution of $x$ because response variable $y$ is unknown in the testing data. In this paper, we propose a novel method, called absolute transfer, that can simultaneously transfer variable $x$ and variable $y$. In fact, absolute transfer considers the joint distribution $p(x,y)$. Compared with previous work, our framework does not require prior information and can efficiently control the bias associated with an unknown response variable $y$ in the testing data.

In this section, we attempt to learn a $(p+1) \times q$ affine matrix $\mathcal{B}$ to transform the data into a new $q$-dimensional feature space. However, absolute transfer must overcome the fact that there is no response variable $y$ in the testing data. As a result, we must learn a method for transferring explanatory variable $x$ for the testing data and transferring explanatory variable $x$ as well as response value $y$ for the training data to the new feature space by minimizing the distance between any two domains, including the latent domains and the target domain from testing data. Note that the affine matrix for the training data and the testing data should be the same to ensure the transformation is consistent. From our perspective, the case in which $y$ is unknown for the target domain is equivalent to setting the values of response variable $y$ to zero. In corresponding method, the response variable $y$ in the training data can be normalized by the z-score and reduced to simplify the transfer of the source latent domains to the target domain. Then, the same affine matrix can be used to transfer explanatory variable $x$ for the testing data and to transfer explanatory variable $x$ as well as response variable $y$ for the training data. 

However, if we assign zero to the values of response variable $y$ in the testing data, we would introduce an average bias to the model. The current solution introduces this problem. We provide two solutions:

1. Our framework takes parameter $\alpha \textless 1$ to reduce the normalized response variable $y$ for the training data. In this way, reducing the response variable $y$ for the training data would help to simultaneously control the bias and to facilitate transfer.

2. We also try to regularize the affine matrix for response variable $y$, as introduced in the PCA with Response Regularization section. This approach would also help to control the bias. 

In summary, for response variable $y$, we let:
$$ \hat{y}=\left\{
\begin{aligned}
& \alpha\overline{y}, & y \; is \; training \; data \; and \; \overline{y} \; is \; calculated \; by \; zscore\\
& 0. & y \; is \; testing \; data
\end{aligned}
\right.
$$
In this paper, we let $d_i=(x_i,\hat{y_i})$, $d^k=\{d_1^k, ..., d_{n_k}^k\}$ and $d^l=\{d_1^l, ..., d_{n_l}^l\}$ be two sets of instances from the $k$-th and $l$-th domains respectively, and we let $\Psi$ be the transform function. Then, the distance between the $k$-th and $l$-th domains in the transformed feature space can be expressed as \begin{equation}dist(d^k,d^l)=||(\frac{1}{n_k}\sum_{i=1}^{n_k}\Psi(d_i^k)-\frac{1}{n_l}\sum_{j=1}^{n_l}\Psi(d_j^l)||^2.\end{equation} If we let $D^{kl}=[d^k,d^l]^T$, then the equation can be simplified to \begin{equation}dist(d^k,d^l)=tr(\Psi(D^{kl})S^{kl}\Psi^T(D^{kl})),\end{equation} and $S^{kl}$ is the MMD matrix computed as follows:
$$S_{ij}^{kl}=\left\{
\begin{aligned}
& \frac{1}{n_k^2}, & d_i,d_j \in d^k \\
&\frac{1}{n_l^2}, & d_i,d_j \in d^l \\
& -\frac{1}{n_kn_l}. & otherwise
\end{aligned}
\right.
$$
Transfer learning attempts to minimize this distance. However, in this paper, the latent domain information is mined by the Dirichlet process, and multiple domains must be considered. In the proposed scheme, the distances of all instances in all domains, including $m$ latent domains and the target domain, should be as small as possible. Then, the distances of all $m+1$ domains can be expressed as follows: \begin{equation}DIST=\sum_{k>l}dist(d^k,d^l).\end{equation}

In summary, suppose there are $m+1$ domains including $m$ mined latent domains and the target domain from the testing data. Then, like ~\cite{ref_article7}, the distances can be simplified as follows: \begin{equation}DIST=tr(\Psi(D)S\Psi^T(D)),\end{equation} 
where $D=[d_1^1, ..., d_{n_1}^1, ..., d_1^m, ..., d_{n_m}^m, d_1^{m+1}, ..., d_{n_{m+1}}^{m+1}]^T$, including all the latent domains and the target domain, and $S$ is defined as
$$ S_{ij}=\left\{
\begin{aligned}
& \frac{m}{n_k^2}, & d_i,d_j \in d^k \\
& -\frac{1}{n_kn_l}. & d_i \in d^k, d_j \in d^l
\end{aligned}
\right.
$$
As shown above, as the output of Eq.(11) decreases, the distributions of any two domains become closer.

To summarize, we let variable $D$ be \begin{equation}D=(x,\hat{y})^T,\end{equation} which consists of all the latent domains and the target domain. The distance of domains in the transformed space can be expressed as
\begin{align}
DIST & = tr(\mathcal{B}^TDSD^T\mathcal{B}).
\end{align}

\subsection{Regularization}
Given the latent domain information that is estimated by the Dirichlet process, minimizing the MMD distance should not destroy much of the original information. To solve this problem, previous works consider several regularization methods. In this paper, we integrate two major regularization schemas and our proposed the regularization of response variable $y$ into the framework, as introduced previously. 

\subsubsection{Graph Regularization}
For our automatic transfer learning method based on a Dirichlet process, our feature mapping should be smooth with respect to the intrinsic manifold structure of the original data. Graph regularization attempts to find a feature mapping $\psi(x)$. Let the Laplacian matrix $L$ be the normalized graph Laplacian, $\mathcal{D}$ be the diagonal degree matrix of $L$ and $W$ be the adjacency matrix, as introduced in~\cite{ref_article30}. The Laplace--Beltrami operator can be approximated as follows:

\begin{align}
\Omega(\psi) & = \frac{1}{2}\sum_{i,j=1}^{n}{W_{ij}||\frac{\psi(d_i)}{\sqrt{\mathcal{D}_{ii}}} - \frac{\psi(d_j)}{\sqrt{\mathcal{D}_{jj}}}||_2^2}\\
& = tr(\mathcal{B}^TDLD^T\mathcal{B}).
\end{align}

Our automatic transfer learning method based on a Dirichlet process with graph regularization searches for an affine matrix $\mathcal{B}$ such that the instances mapped into the feature space are smooth with respect to the manifold for each domain and the distances for different domains are as small as possible. Like ~\cite{ref_article7}, The process can be formulated as the following optimization problem:

$$
\begin{aligned}
& \min{tr(\mathcal{B}^TDSD^T\mathcal{B})+\tau tr(\mathcal{B}^TDLD^T\mathcal{B})},
\end{aligned}
$$
where parameter $\tau$ controls the importance of graph regularization.

Let $M = S + \tau L$. Then, the formula becomes
$$
\begin{aligned}
& \min{tr(\mathcal{B}^TDMD^T\mathcal{B})}. 
\end{aligned}
$$

\subsubsection{PCA with Response Regularization}
In addition to graph regularization, our automatic transfer learning method based on a Dirichlet process learns a transformed feature representation by minimizing the reconstruction error of the input data. PCA regularization extends the nonparametric MMD to measure the difference in domain distributions. PCA regularization attempts to construct a feature representation that is effective and robust to substantial distributional differences. Denote $H$ as the centring matrix. The learning goal of PCA regularization is to find an orthogonal affine matrix $\mathcal{B}$ to maximize the transformed data variances \begin{equation}\max_{\mathcal{B}^T\mathcal{B}=I}tr(\mathcal{B}^TDHD^T\mathcal{B}).\end{equation} Note that the generalized Rayleigh quotient shows minimizing an equation such that the equation is maximized is equivalent to minimizing an equation such that the equation is fixed. Thus, integrating PCA regularization into our automatic transfer learning framework can lead to the following optimization problem: \begin{equation}\min_{\mathcal{B}^TDHD^T\mathcal{B}=I} tr(\mathcal{B}^TDMD^T\mathcal{B}).\end{equation} 

Furthermore, as introduced previously, we need to regularize the affine matrix for the response variable $y$ to reduce the average bias in our model. Here, we introduce the regularization factor $\beta$. Let \begin{equation}J=diag(1,1,...,1,\beta)\end{equation} be a $(p+1)\times(p+1)$ matrix. Then, by integrating PCA with response regularization, our framework can finally be formulated as \begin{equation}\min_{\mathcal{B}TDHD^T\mathcal{B}=I} tr(\mathcal{B}^TDMD^T\mathcal{B}) + \mu tr(\mathcal{B}^TJ\mathcal{B}),\end{equation} where $\mu$ is the regularization parameter that guarantees that the optimization problem is well-defined.

\subsection{Learning Algorithm}
According to the constrained optimization theory, the following result can be obtained via generalized eigen decomposition \begin{equation}(DMD^T + \mu J)\mathcal{B}=DHD^T\mathcal{B}\Phi.\end{equation} Then, our affine matrix $\mathcal{B}$ can be reduced to solve the equation for the top $q$ eigenvectors.

For our automatic transfer learning method based on a Dirichlet process integrated with graph regularization, PCA and response regularization, graph regularization makes the mapping smooth, PCA regularization makes the mapping robust, and our proposed response regularization alleviates the average bias of our model. Note that the parameter $\tau$ controls the importance of graph regularization. If we set parameters $\tau=0$ and $\beta=1$, the regularization of our framework is reduced to TCA.

The complete procedure of our automatic cross-domain transfer learning is summarized in Algorithm 1.

	
	


\SetArgSty{textnormal}
\begin{algorithm}[h!]
	$\text{\bf{Require:}}$ $X$,$Y$,$a_0$,$b_0$,$a_v$,$b_v$,$a_i$,$b_i$ \\
	
	Initialize variable $\Lambda$, $\Sigma$ and $\nu$\; 
	\While{Not Converge}{
		Update $\mathcal{A}_i$ from $[\mathcal{A}_i|\mathcal{A}_{-i}, \Lambda, \Sigma, \nu, Y, X]$ for $i = 1,...,n$\; 
		Update $Q_k$ from $[Q_k|w, \mathcal{A}, \Lambda, \Sigma, \nu, Y, X]$ for $k = 1,...,m$\; 
		Update $\Sigma^{-1}$ from $[\Sigma^{-1}|Y,X,\mathcal{A},a_0,b_0]$\; 
		Update $\lambda_i^{-1}$ from $[\lambda_i^{-1}|r_i^{k},\Sigma,a_i,b_i]$ for $k = 1,...,m$ and $i = 1,...,p+1$\; 
		Update $\nu$ from $[\nu|a_v,b_v,m]$\; 
	}
	
	\textbf{return} $Q$, $w$ indicates the index of the cluster

	\label{headless_algo}
	\caption{Dirichlet Process for Latent Domain Mining}
\end{algorithm}

\SetArgSty{textnormal}
\begin{algorithm}[h!]
	$\text{\bf{Require:}}$ $\alpha$, $\beta$, $\mu$, $\tau$, $X_{tr}$, $Y_{tr}$, $X_{te}$,$a_0$,$b_0$,$a_v$,$b_v$,$a_i$,$b_i$ \\
	
	Normalize the explanatory value $x$ using the z-score\; 
	Calculate $\hat{y}$ for the response variable $y$\; 
	Learn the latent domains via Dirichlet Process according to Algorithm 2 using $X_{tr}$, $Y_{tr}$, $a_0$, $b_0$, $a_v$, $b_v$, $a_i$, and $b_i$\; 
	Learn the transformed feature space according to formula 19 using $\beta$, $\mu$, and $\tau$\; 
	Learn the linear regression model in the transformed feature space\; 
	Predict the linear regression value in the transformed feature space\; 
	
	\textbf{return} $Y_{te}$

	\label{headless_alg}
	\caption{Automatic Cross-Domain Transfer Learning}
\end{algorithm}

\section{Experiments}
In this section, we conduct numerical experiments for analysing the performance of our proposed automatic cross-domain transfer learning method for linear regression.

Our analysis was implemented on several real datasets (i.e., the forest fires, student, slump test, Istanbul stock and air foil datasets), which are all regression problems. The forest fire dataset tries to predict the burned area of forest fires in the northeast region of Portugal. The student dataset considers student achievements in secondary education at two Portuguese schools. The slump test dataset tries to predict the slump of concrete, which is determined not only by the water content but also by other concrete ingredients. The Istanbul stock dataset includes returns of the Istanbul Stock Exchange and other international indexes; for this dataset, we conduct two experiments in which the stock is traded in TL or USD. The air foil dataset consists of NACA 0012 air foils of different sizes at various wind tunnel speeds and angles of attack. All the datasets are taken from the UCI, and all of them are complex regression problems that are influenced by many factors. Thus, it is appropriate to mine latent domains for these scenarios, and our proposed automatic cross-domain transfer learning method for linear regression achieves outstanding performance.

We compare our method to some state-of-the-art transfer learning methods, including base ridge regression (RR). Since the domains are latent, traditional transfer learning methods take training data as the source domain and testing data as the target domain. 

The regression performance is measured in terms of the root mean square error (RMSE). We use RMSE to measure the performance because $p(y|x)$ follows a normal distribution. The objective function of our latent domain mining and prediction is the MSE, and we use RMSE to unify the dimensions.

Table 1 shows the RMSE for all the datasets. For the table, ACDT is our automatic cross-domain transfer learning. RR represents the original ridge regression. LAPLACE is a model-level semi-supervised (transductive) learning algorithm. Maximum independence domain adaptation (MIDA) is a feature-level transfer learning (domain adaptation) algorithm that augments the features and then learns a domain-invariant subspace. TCA is transfer learning integrated with PCA regularization, and JDA is TCA for the joint distribution $p(x,y)$ with pseudo-response variable $y$ for the testing data. From the results, we find that transfer learning based on latent domains mined by a Dirichlet process performs much better than the other transfer learning methods in all the senarios. As demonstrated previously, our framework performs better because we mine the efficient latent domains, our transfer learning method considers the joint distribution $p(x,y)$ that contains more useful information, and we control the bias efficiently with proper regularization.

\begin{table}[h]
	\centering
	\caption{RMSE}
	\footnotesize
	\begin{tabular}{ccccccc}
		\hline
		& RR& LAPLACE& MIDA& TCA& JDA& ACDT\\
		\hline
		forest& 0.4650& 0.4580& 0.3775& 0.3946& 0.3876& {\bfseries 0.3761}\\
		student& 0.8091& 0.7745& 0.8006& 0.7585& 0.7738& {\bfseries 0.7542}\\
		slump& 0.3666& 0.7087& 1.1154& 0.3665& 0.3597& {\bfseries 0.3556}\\
		stockTL& 0.6717& 1.0481& 1.0492& 0.9429& 0.6710& {\bfseries 0.6703}\\
		stockUSD& 0.6793& 0.8048& 0.8601& 0.6691& 0.6713& {\bfseries 0.6506}\\
		airfoil& 0.7067& 0.7577& 1.0334& 0.7067& 0.7068& {\bfseries 0.7039}\\
		\hline
	\end{tabular}
\end{table}

\begin{figure*}[htbp]
	\centering
	\includegraphics[width=4.1cm,height=3.5cm]{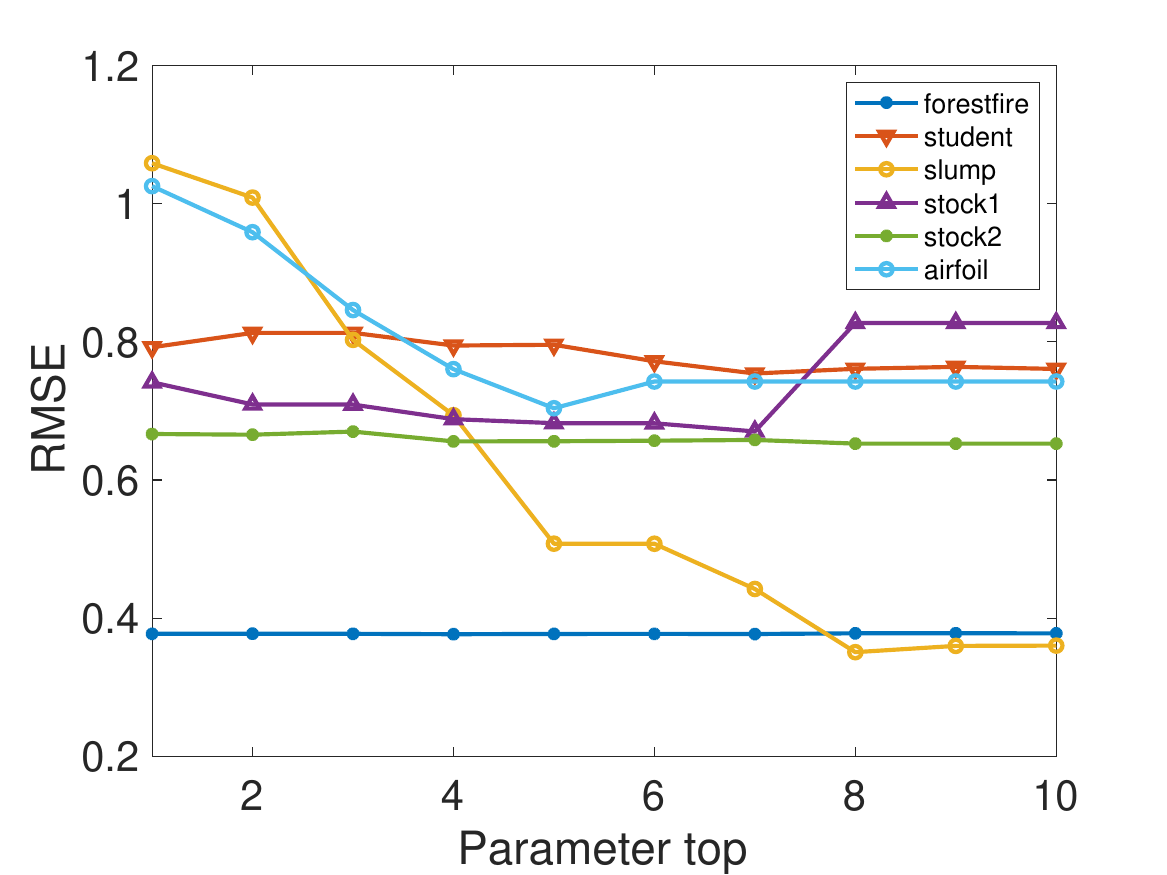}
	\includegraphics[width=4.1cm,height=3.5cm]{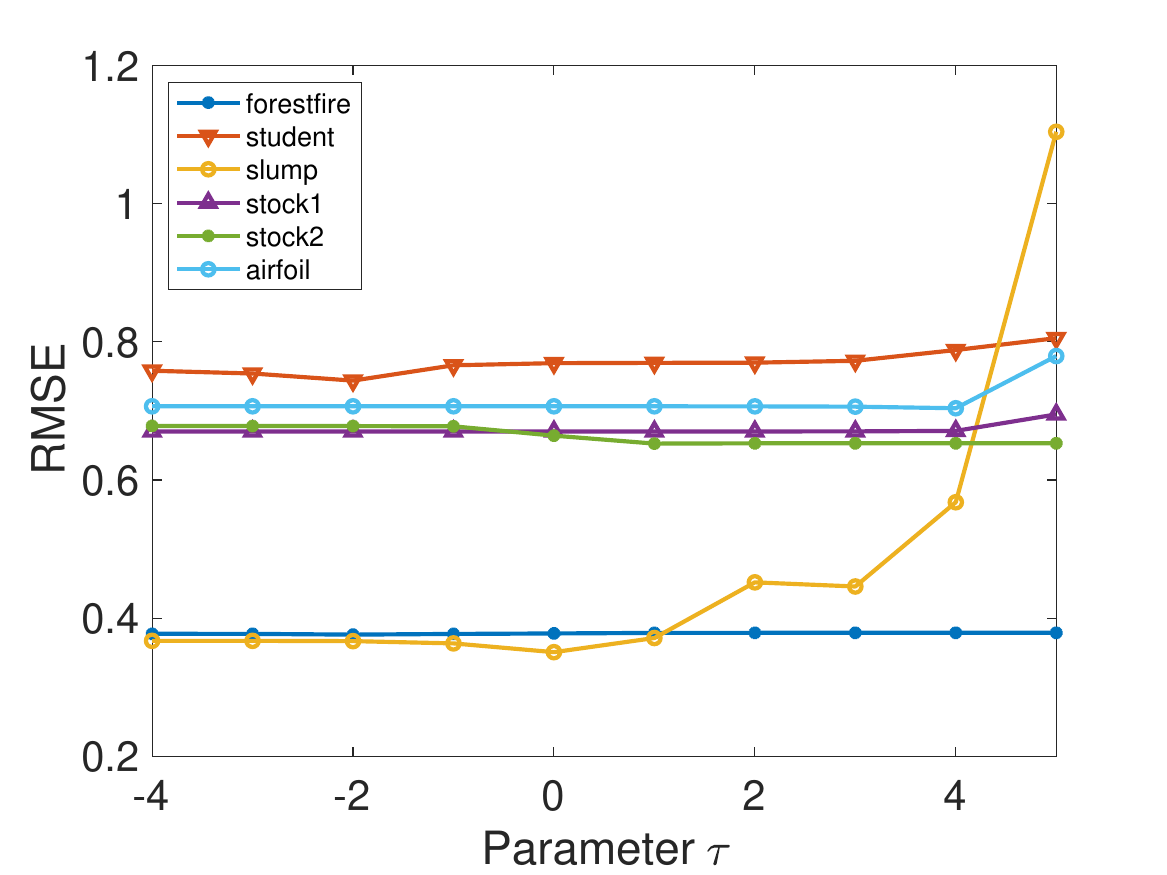}
	\includegraphics[width=4.1cm,height=3.5cm]{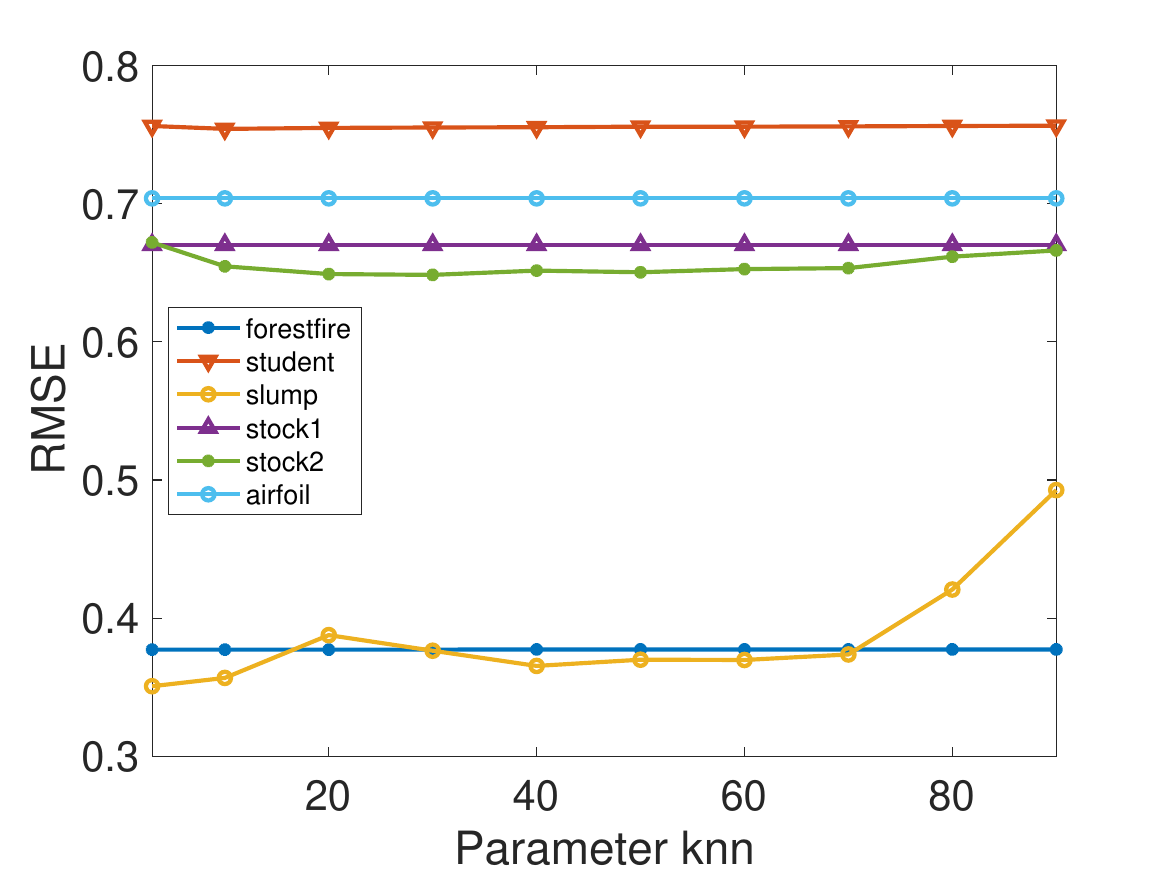}
	\includegraphics[width=4.1cm,height=3.5cm]{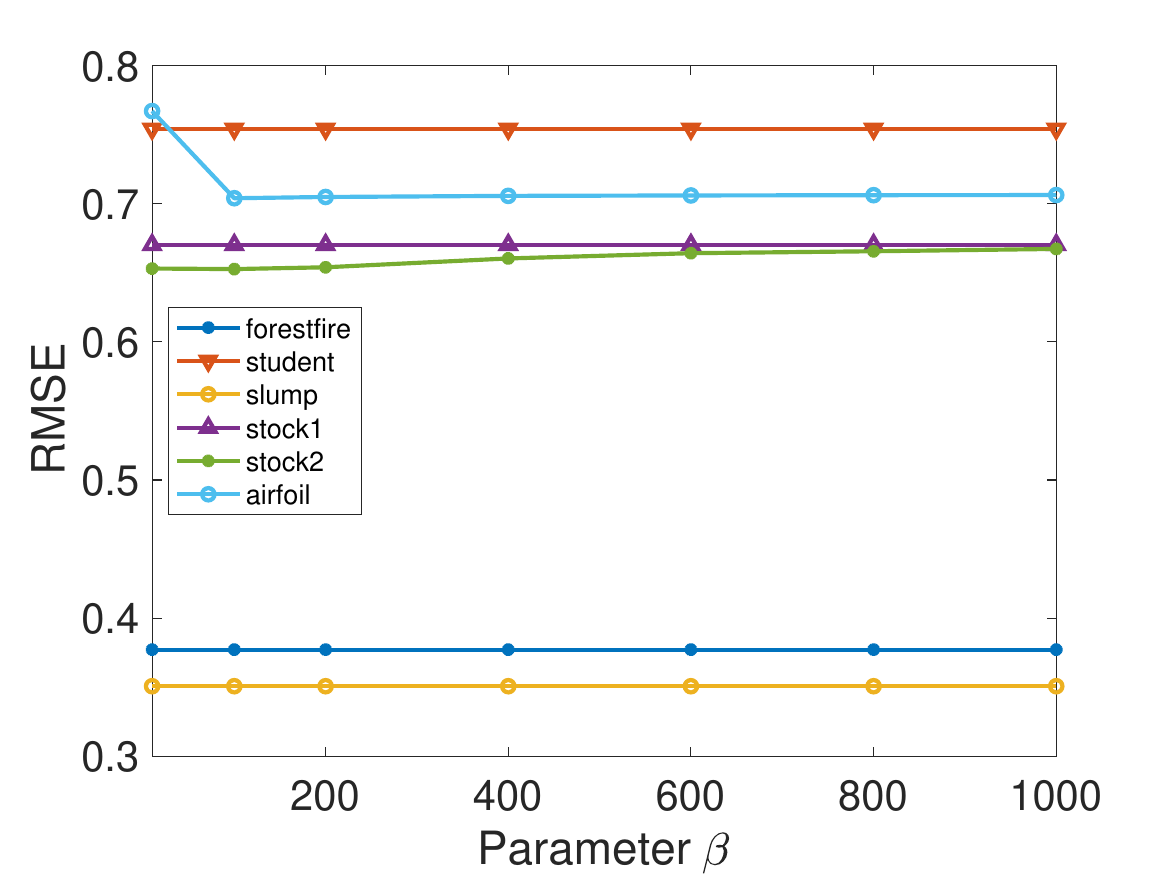}
	\caption{Top feature, $\tau$ for graph regularization, knn and $\beta$ for regularization of the response value influence on the performance of our automatic cross-domain transfer learning for linear regression.}
	\label{fig3}
\end{figure*}

Figure 2 shows the top feature, $\tau$ for graph regularization, knn and $\beta$ for regularization of the response value to influence the performance of our automatic cross-domain transfer learning method for linear regression. From the results, we can see that our framework is not sensitive to most parameters in these datasets except for the slump dataset. Thus, our model is robust enough for real applications. It would be easy to tune the parameters for the transfer process in our framework.

Furthermore, for the latent domain mining process in our framework, we show the parameters influence the number of latent domains mined by Dirichlet process for each dataset in Figure 3. The number of clusters is sensitive to the parameters: the parameters should be fine-tuned to make the mined latent domains converge. However, the three relevant parameters can be searched heuristically within 10 to 100 for real normalized data. Additionally, from a heuristic perspective, a large value of $a_0$ or a small value of $a_v$ help our model to converge. Typically, hyperparameter $a_v$ controls the concentration parameter $\nu$ of the Dirichlet process prior. A small value of $a_v$ results in a small value of $\nu$. Then, we would obtain fewer clusters in the Dirichlet process. 

\begin{figure}[!htb]
	\centering
	\includegraphics[width=4.1cm,height=3.5cm]{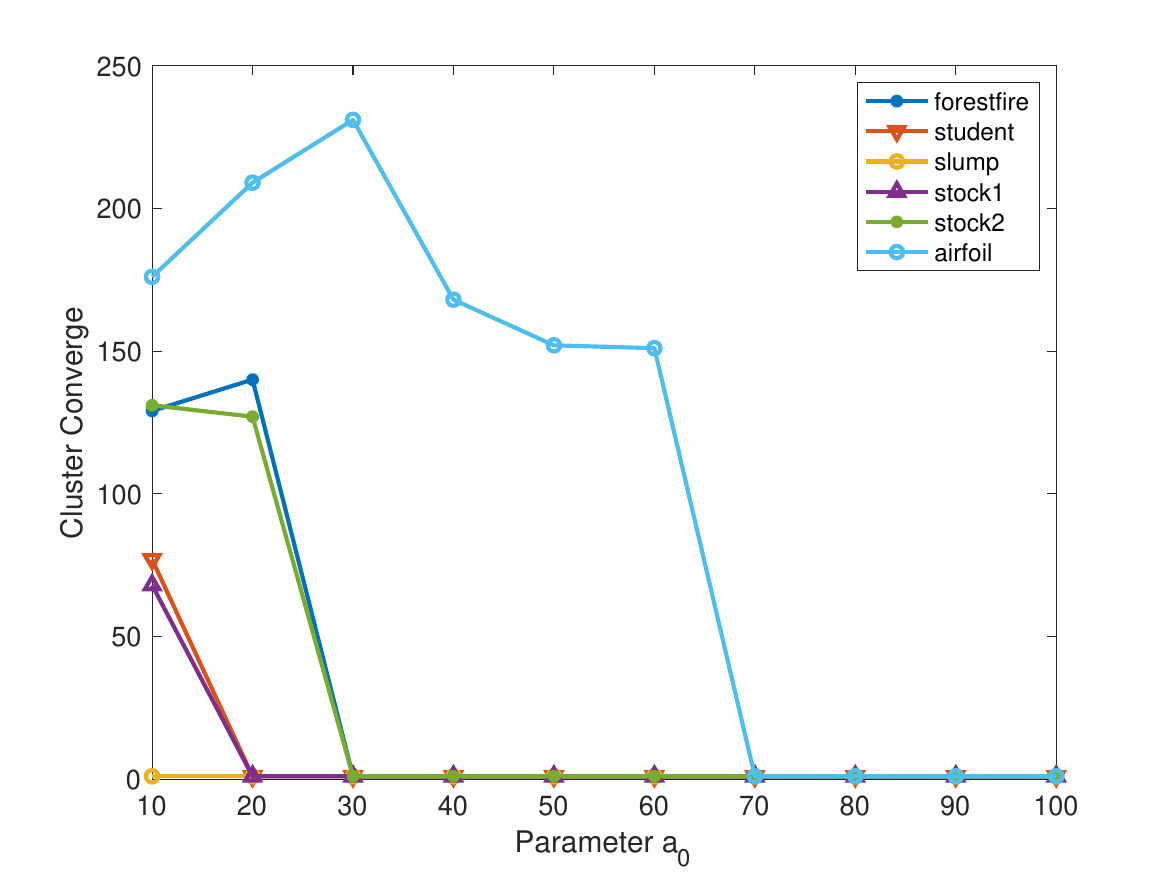}
	\includegraphics[width=4.1cm,height=3.5cm]{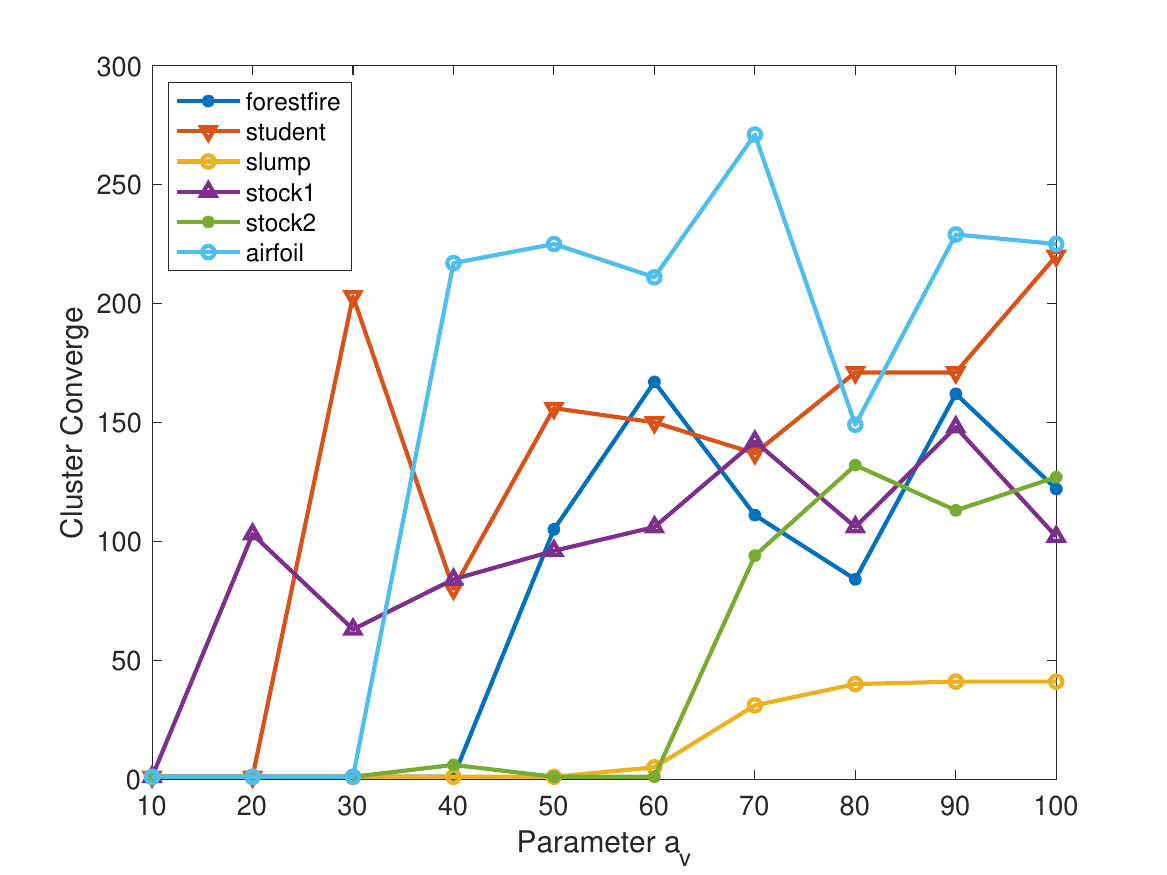}
	\includegraphics[width=4.1cm,height=3.5cm]{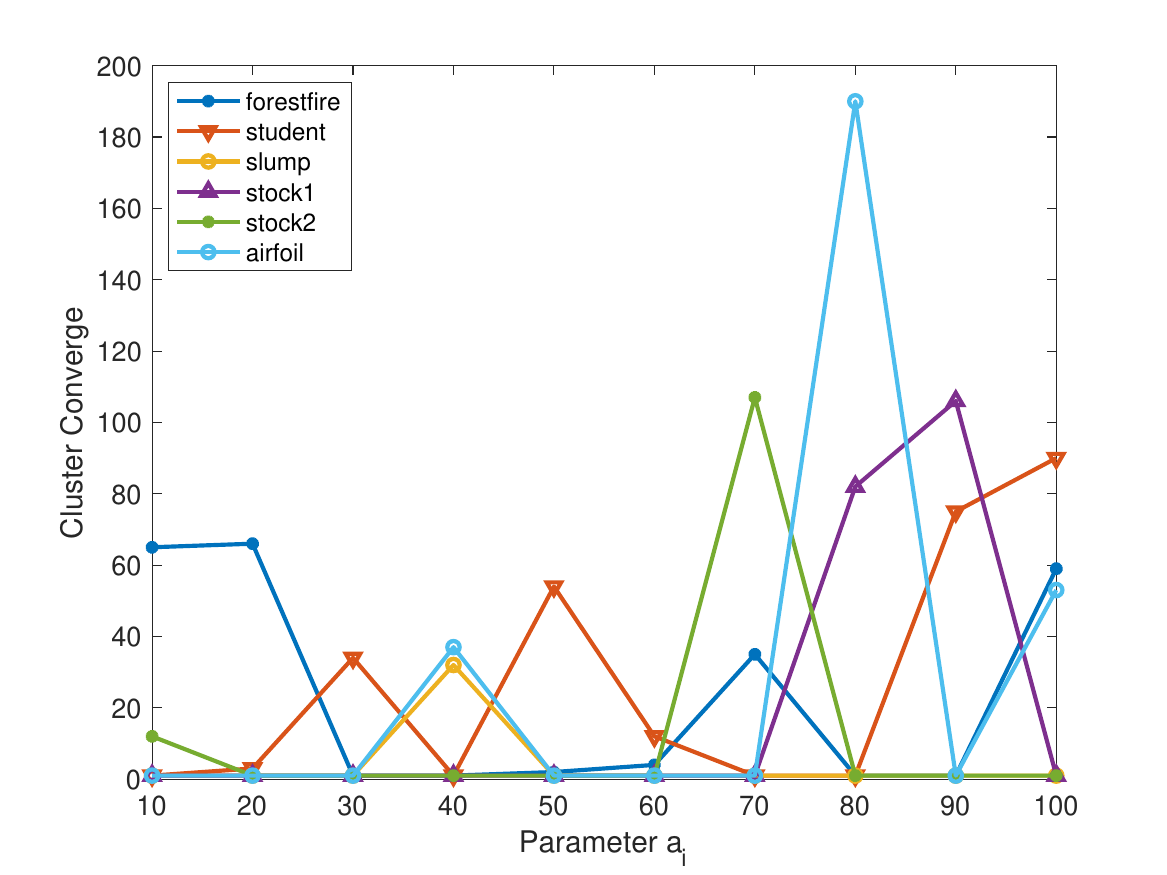}
	\caption{The number of latent domains mined by Dirichlet process on hyperparameters $a_0$, $a_v$ and $a_i$ for each dataset.}
	\label{fig3}
\end{figure}

Next to show the clustering property of latent domains for effective transfer learning, in Figure 4, we present one image for each dataset. In the figure, for each latent domain, we compress the original data to two dimensions via PCA, and we can see that the instances in each latent domain tend to cluster based on some rule. Furthermore, the distributions of the instances for the latent domains are relatively different; thus, the latent domains mined by our framework are reasonable. As a result, transfer learning can be applied for these latent domains. Note that the clustering property may be not explicit in this view for the student set, as it has been compressed by PCA. However, according to Table 1, the performance of our framework on the student set is excellent compared with state-of-the-art methods.

\begin{figure}[!htb]
	\centering
	\includegraphics[width=3.9cm,height=3cm]{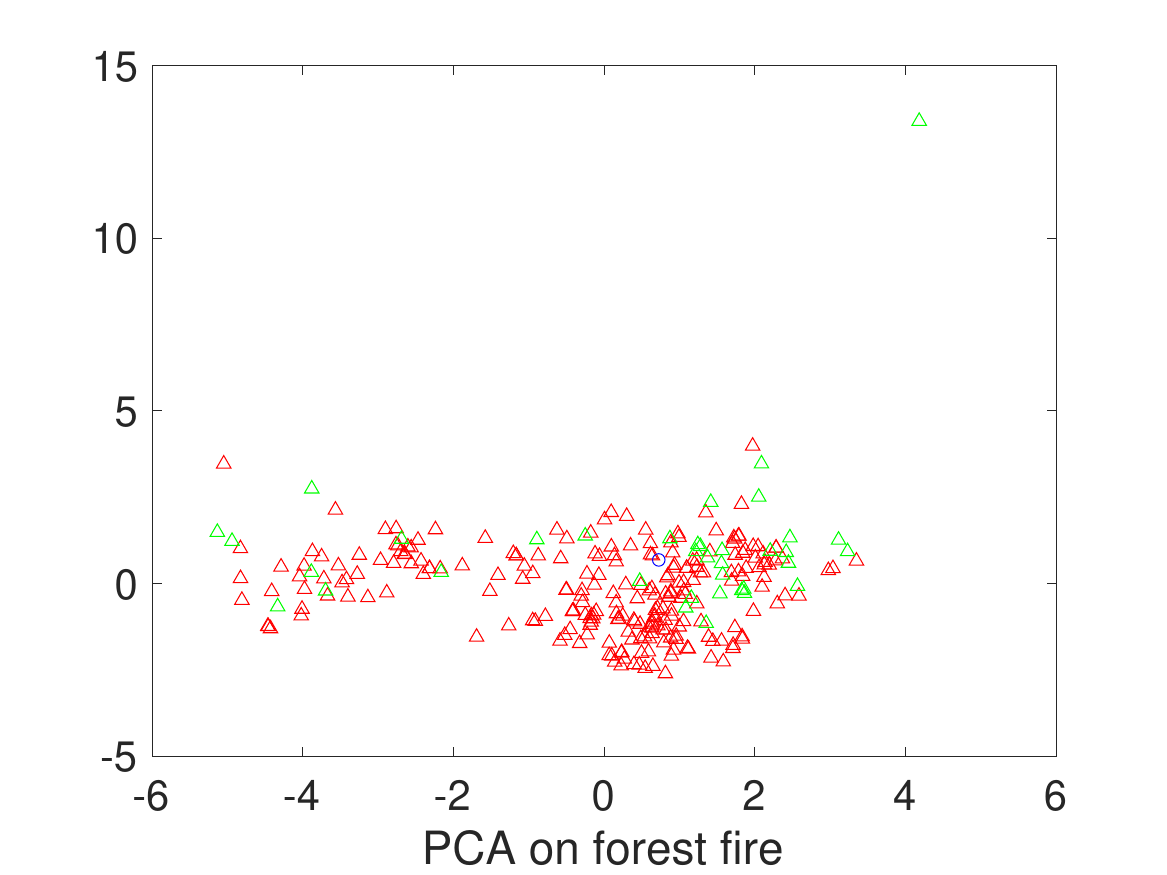}
	\includegraphics[width=3.9cm,height=3cm]{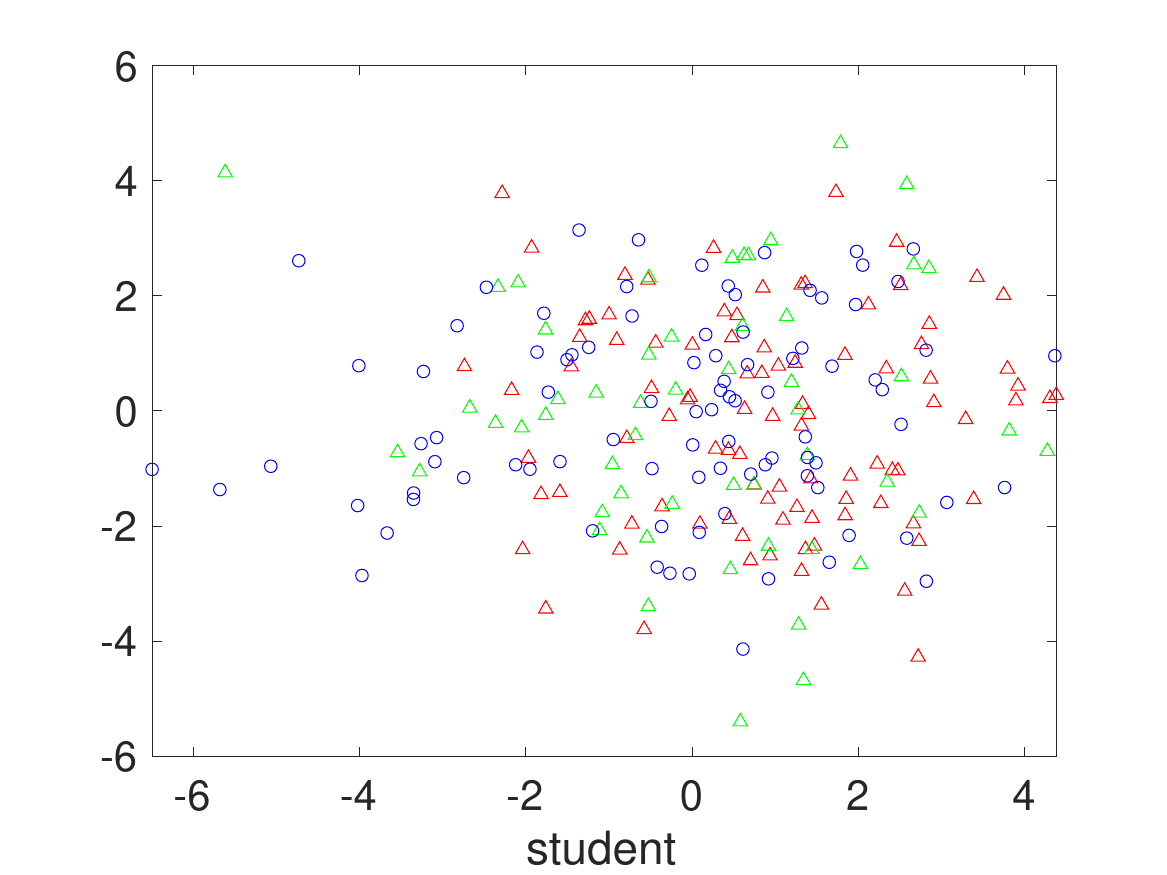}
	\includegraphics[width=3.9cm,height=3cm]{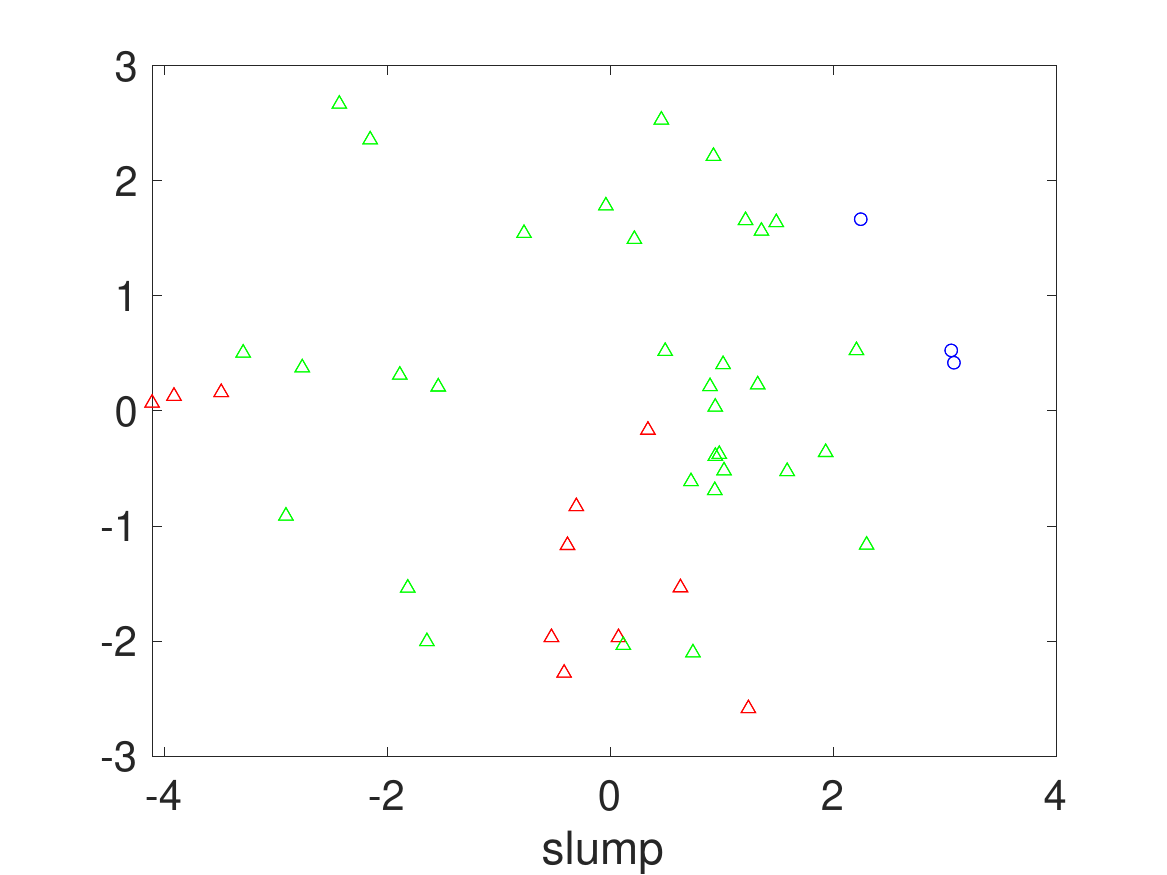}
	\includegraphics[width=3.9cm,height=3cm]{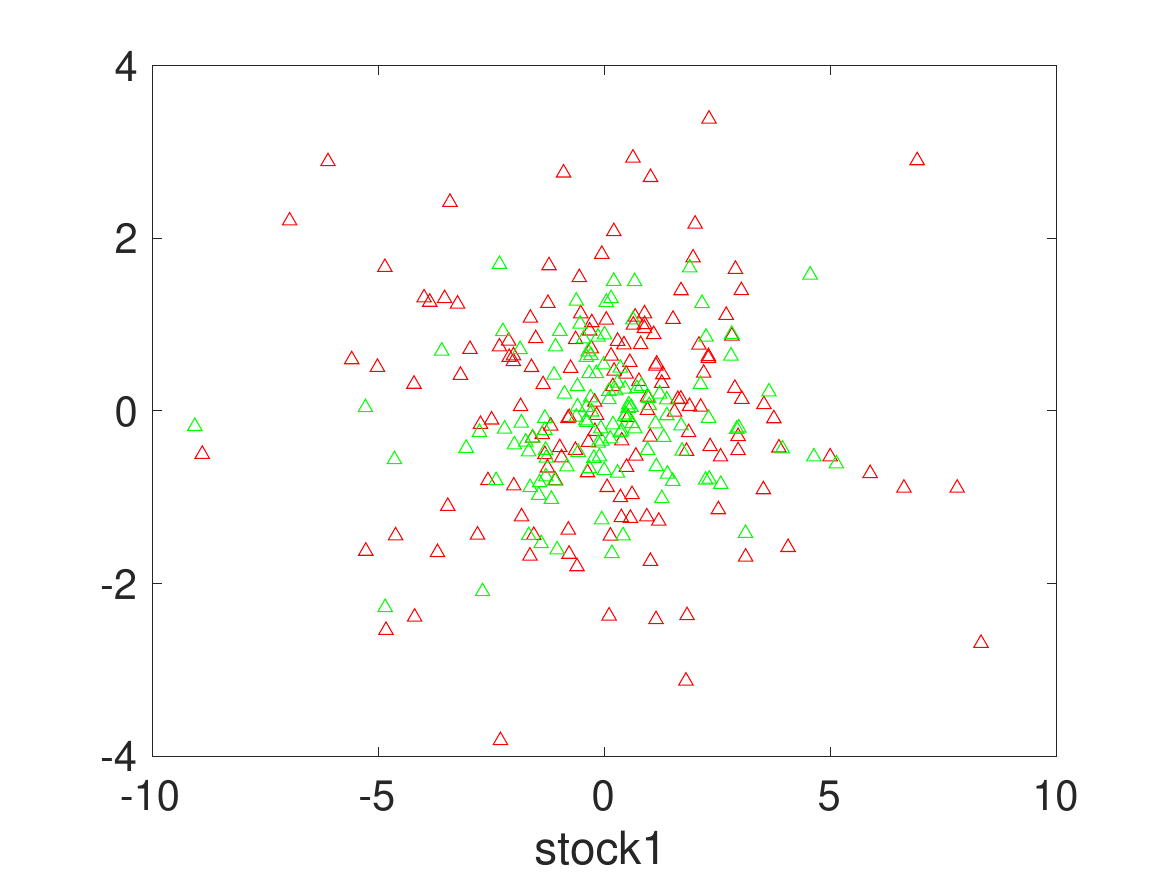}
	\includegraphics[width=3.9cm,height=3cm]{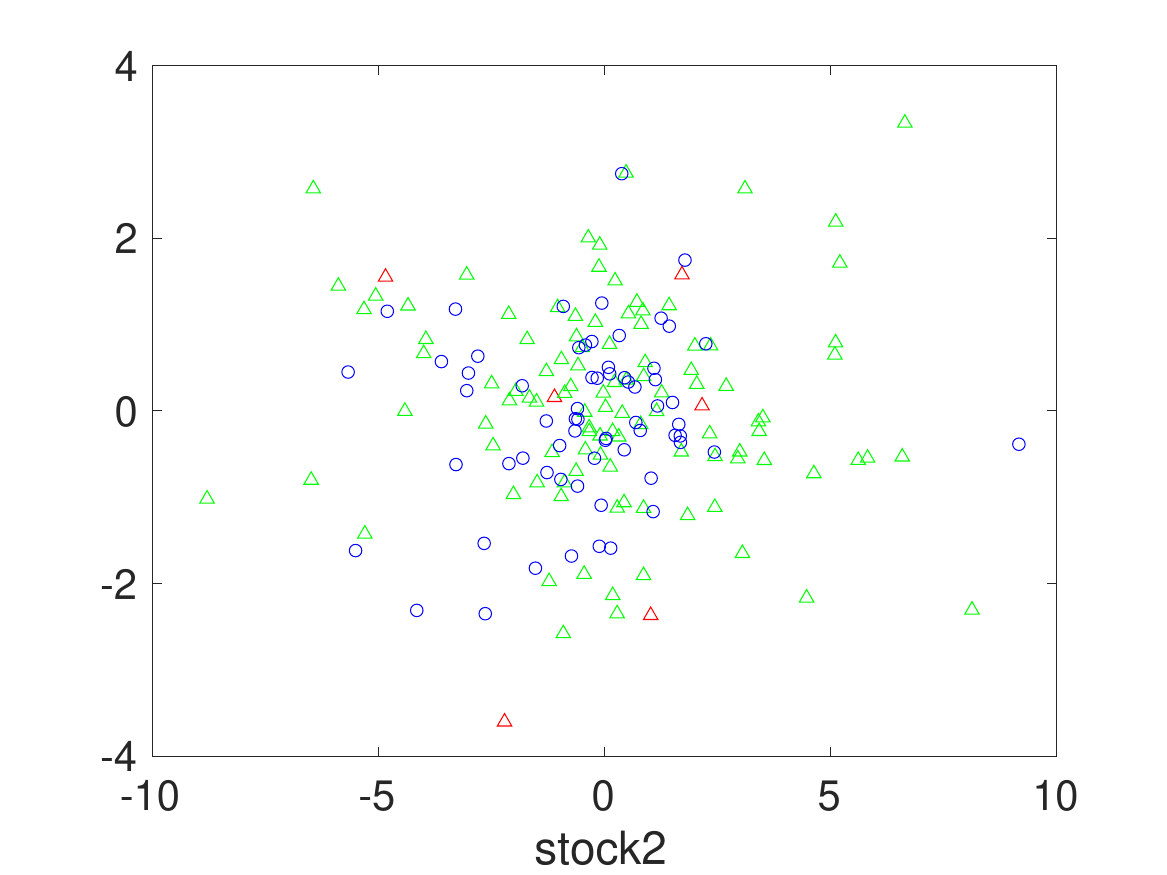}
	\includegraphics[width=3.9cm,height=3cm]{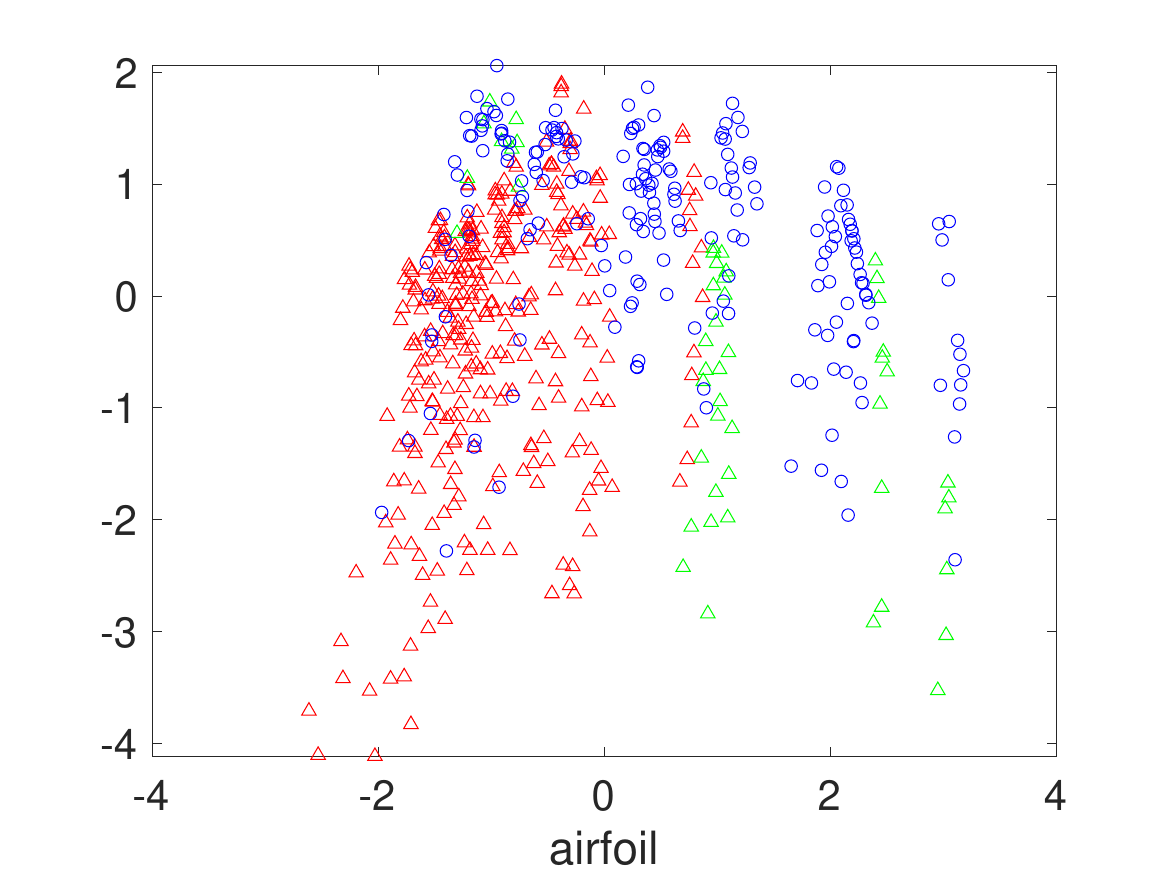}
	\caption{Mined latent domains for each dataset. The two dimensions are obtained via PCA for the original data.}
	\label{fig3}
\end{figure}

Finally, we present the distributions of the instances for each latent domain after transfer learning in Figure 5, which shows the two dimensions calculated by our method. As discussed previously, we mine the latent domain according to the coefficients and conduct transfer learning by the adaptation of $p(x,y)$ to make the distance  any two latent domains and the target domain be as small as possible. Thus, efficient transfer learning results in better performance for our framework achieving. For Figure 4 and Figure 5, we show the top latent domains for each dataset to enhance the explanation.

\begin{figure}[!htb]
	\centering
	\includegraphics[width=3.9cm,height=3cm]{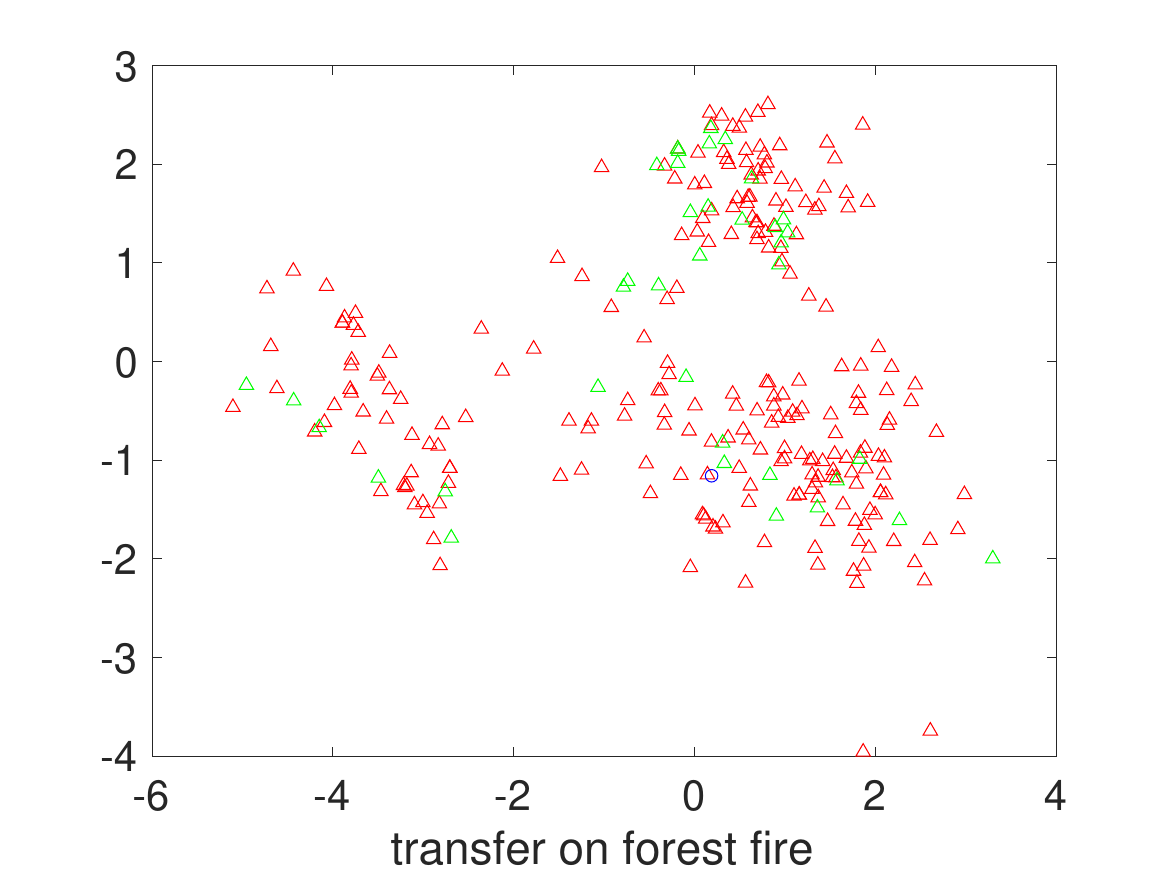}
	\includegraphics[width=3.9cm,height=3cm]{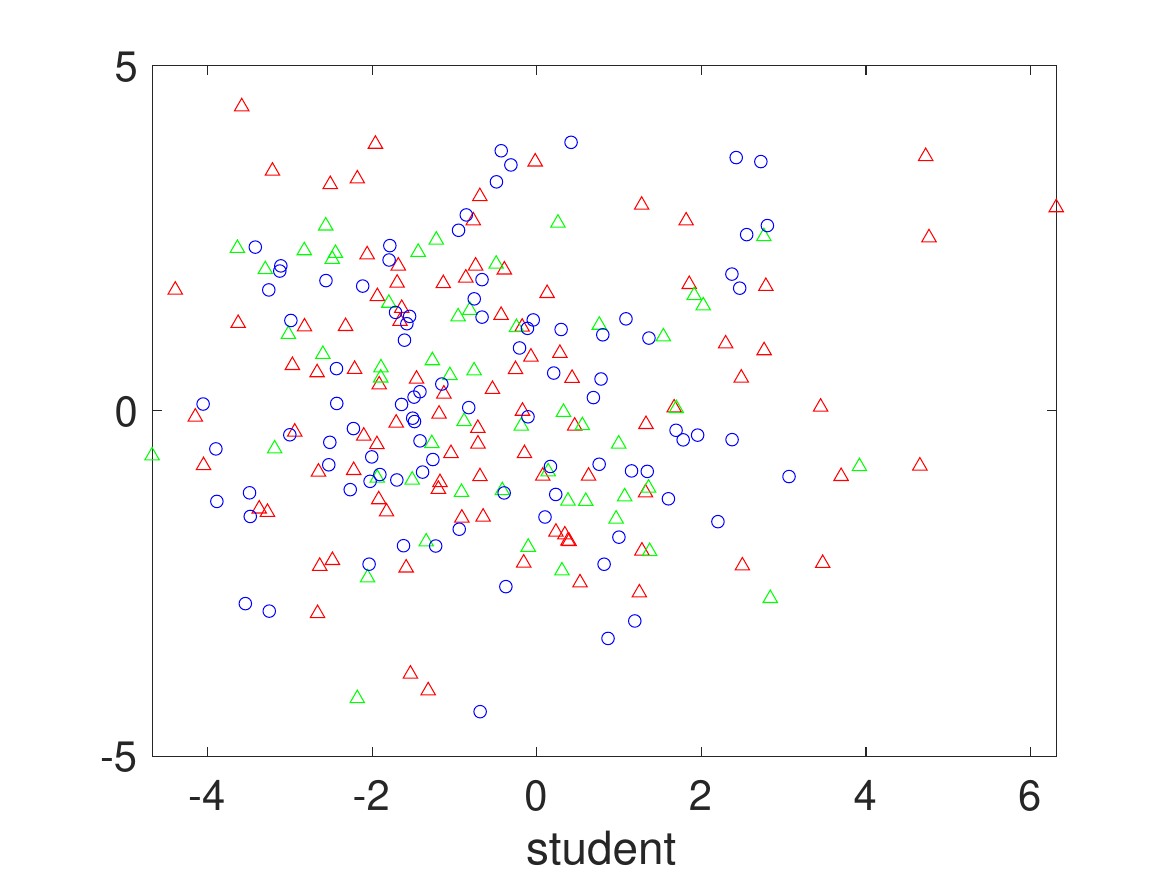}
	\includegraphics[width=3.9cm,height=3cm]{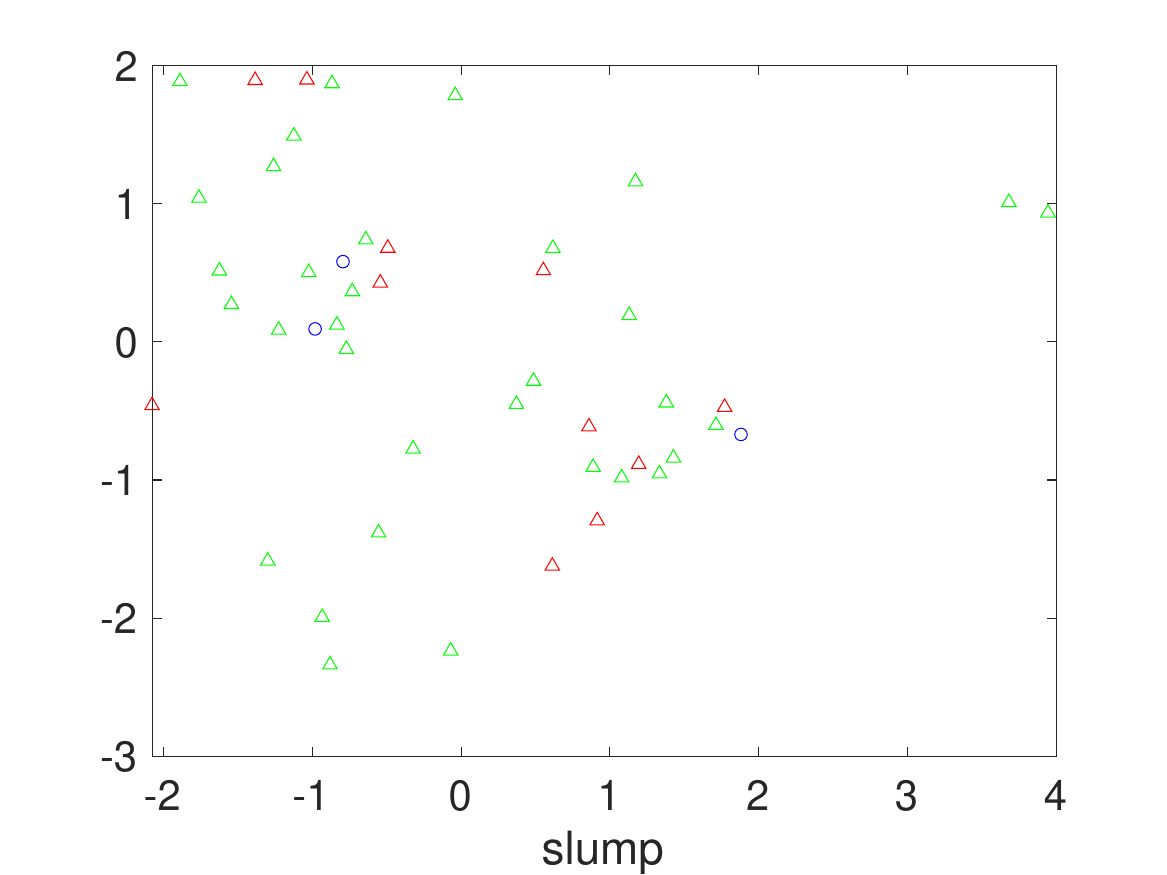}
	\includegraphics[width=3.9cm,height=3cm]{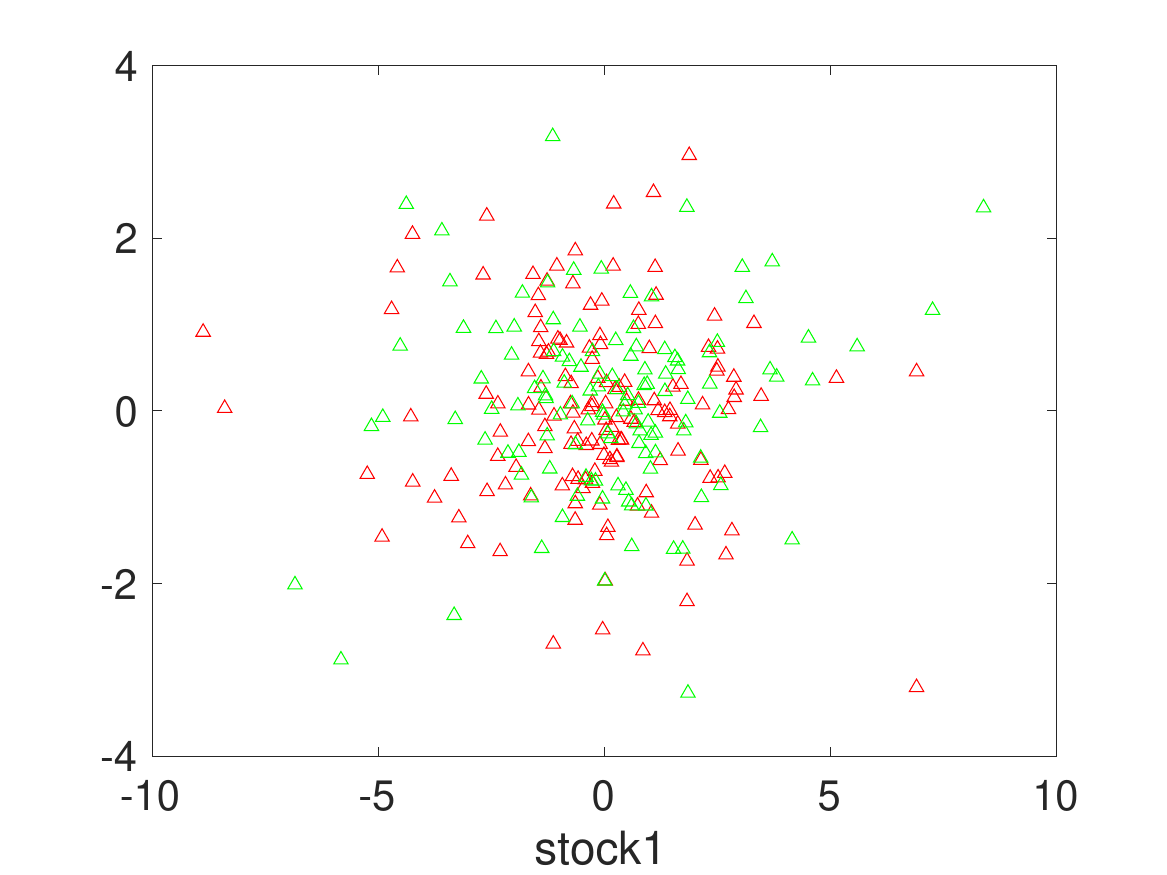}
	\includegraphics[width=3.9cm,height=3cm]{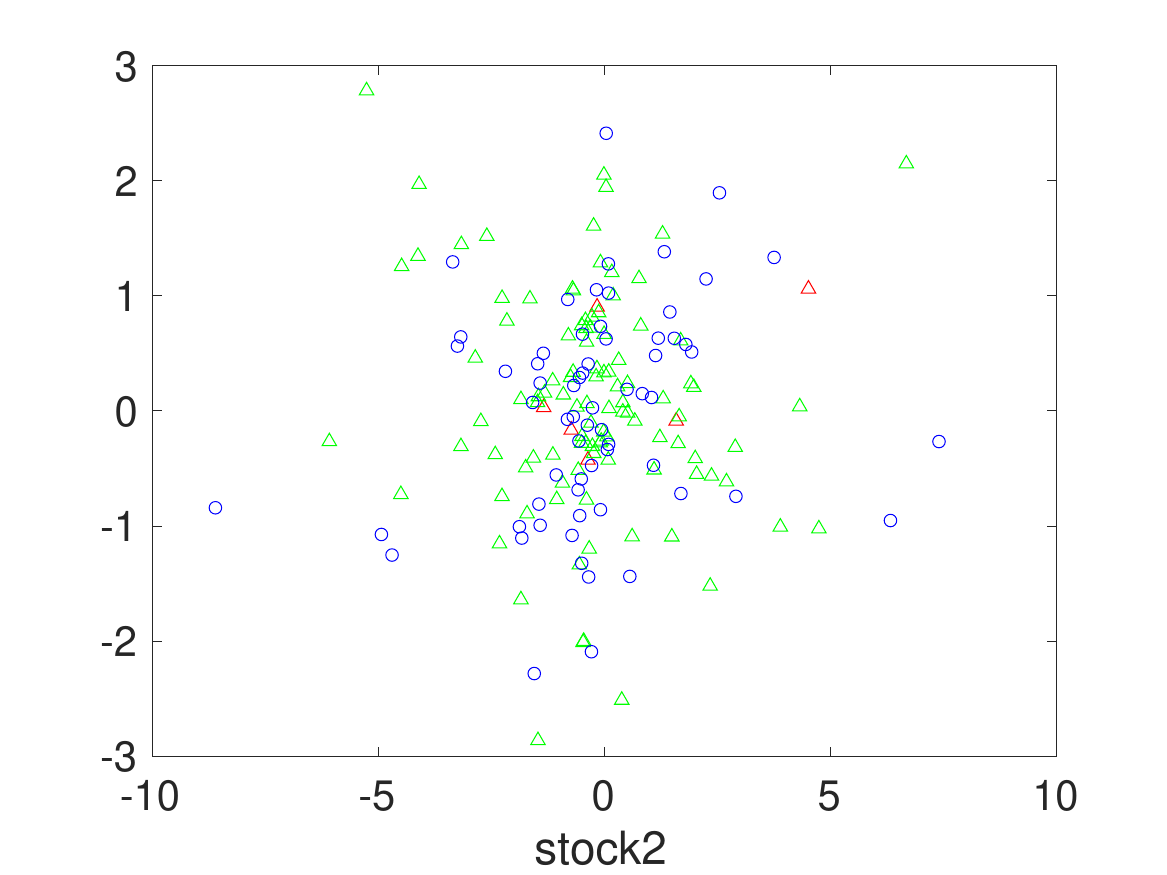}
	\includegraphics[width=3.9cm,height=3cm]{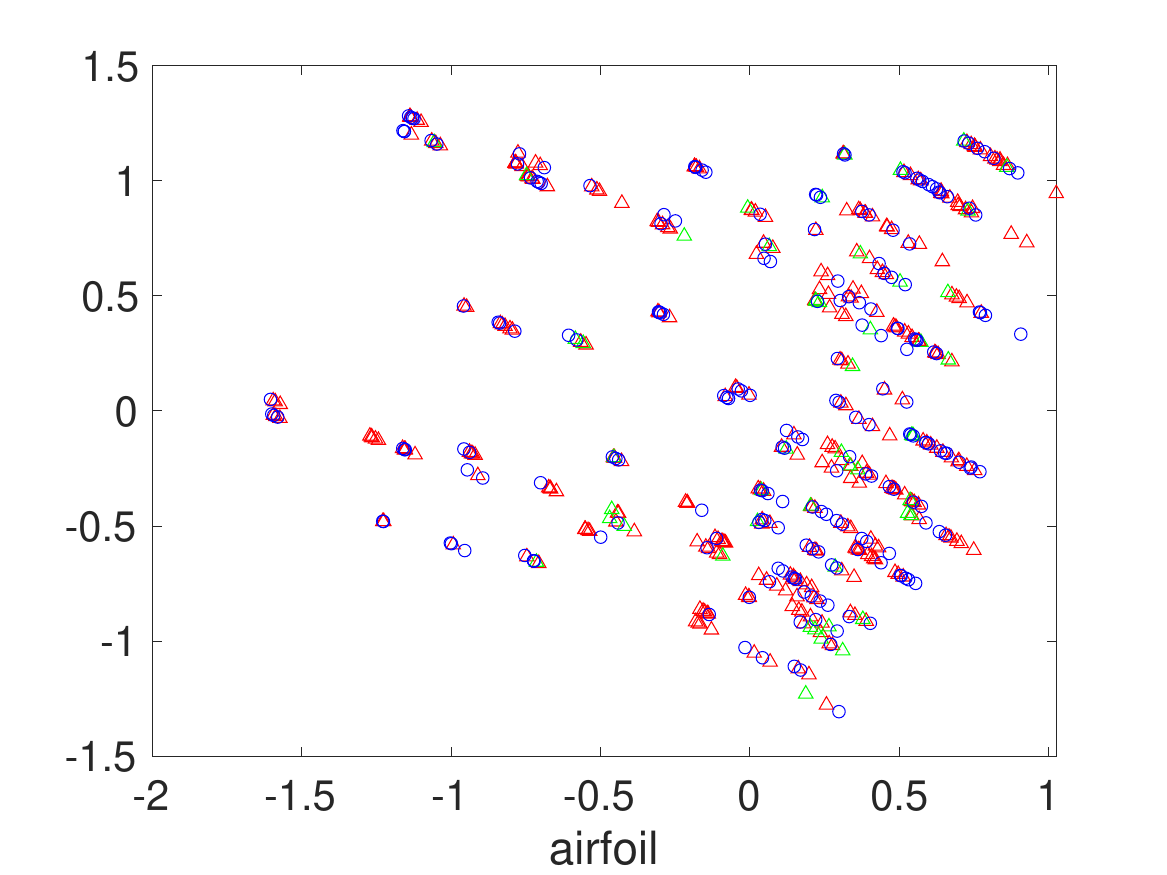}
	\caption{Transformed feature space for each dataset. The two dimensions are obtained via our transferred feature space.}
	\label{fig3}
\end{figure}

\section{Conclusion}
In this paper, we extend transfer learning to the situation in which domain information is uncertain. We introduce our automatic cross-domain transfer learning method for linear regression problems, which can also be extended to classification problems, as the likelihood function can be changed to a Bernoulli distribution or another exponential family distribution for classification. For the real datasets in our experiments, we assume that some latent domain information is available for transfer learning. We obtain this information based on the regression coefficients by a Dirichlet process technique. We then learn a transferred feature space using MMD integrated with PCA regularization, response regularization, and graph regularization to adapt the joint distribution $p(x,y)$ of the latent domains and the target domain. Our novel framework considers the fact that the testing data have no response variable $y$. Compared with previous work, our framework efficiently controls the bias associated with the unknown response variable $y$ for testing data. Few studies have thoroughly analysed this problem. We perform linear regression on the new feature space that minimizes the distributions of different latent domains and the target domain. The experimental results show that the proposed model performs well on real data.

\end{document}